\documentclass[sn-mathphys-num]{sn-jnl}

\usepackage{graphicx}%
\usepackage{multirow}%
\usepackage{amsmath,amssymb,amsfonts}%
\usepackage{amsthm}%
\usepackage{mathrsfs}%
\usepackage[title]{appendix}%
\usepackage{xcolor}%
\usepackage{textcomp}%
\usepackage{manyfoot}%
\usepackage{booktabs}%
\usepackage{algorithm}%
\usepackage{algorithmicx}%
\usepackage{algpseudocode}%
\usepackage{listings}%
\usepackage{makecell}
\usepackage{bbding}
\usepackage{graphicx}
\usepackage{subfigure}
\usepackage{float} 
\usepackage{multirow}  
\usepackage{array}     

\theoremstyle{thmstyleone}%
\newtheorem{theorem}{Theorem}
\theoremstyle{thmstyletwo}%

\theoremstyle{thmstylethree}%
\newtheorem{definition}{Definition}%

\begin{document}

\title[Article Title]{AutoMathKG: The automated mathematical knowledge graph based on LLM and vector database}

\author[1,2]{\fnm{Rong} \sur{Bian}}\email{bianrong2021@amss.ac.cn}

\author[1,2]{\fnm{Yu} \sur{Geng}}\email{gengyu2020@amss.ac.cn}

\author[1,2]{\fnm{Zijian} \sur{Yang}}\email{yangzijian20@mails.ucas.ac.cn}
\author*[1,3,4]{\fnm{Bing} \sur{Cheng}}\email{bc2@amss.ac.cn}

\affil*[1]{\orgdiv{Academy of Mathematics and Systems Science}, \orgname{Chinese Academy of Sciences}, \orgaddress{\city{Beijing}, \postcode{100190}, \country{China}}}
\affil[2]{\orgdiv{School of Mathematical Sciences}, \orgname{University of Chinese Academy of Sciences}, \orgaddress{\city{Beijing}, \postcode{100049}, \country{China}}}
\affil*[3]{\orgdiv{AMSS Center for Forecasting Science}, \orgname{Chinese Academy of Sciences}, \orgaddress{\city{Beijing}, \postcode{100190}, \country{China}}}
\affil*[4]{\orgdiv{State Key Laboratory of Mathematical Science, Academy of Mathematics and Systems Science}, \orgname{Chinese Academy of Sciences}, \orgaddress{\city{Beijing}, \postcode{100190}, \country{China}}}


\abstract{A mathematical knowledge graph (KG) presents knowledge within the field of mathematics in a structured manner. Constructing a math KG using natural language is an essential but challenging task. There are two major limitations of existing works: first, they are constrained by corpus completeness, often discarding or manually supplementing incomplete knowledge; second, they typically fail to fully automate the integration of diverse knowledge sources. This paper proposes AutoMathKG, a high-quality, wide-coverage, and multi-dimensional math KG capable of automatic updates. AutoMathKG regards mathematics as a vast directed graph composed of Definition, Theorem, and Problem entities, with their reference relationships as edges. It integrates knowledge from ProofWiki, textbooks, arXiv papers, and TheoremQA, enhancing entities and relationships with large language models (LLMs) via in-context learning for data augmentation. To search for similar entities, MathVD, a vector database, is built through two designed embedding strategies using SBERT. To automatically update, two mechanisms are proposed. For knowledge completion mechanism, Math LLM is developed to interact with AutoMathKG, providing missing proofs or solutions. For knowledge fusion mechanism, MathVD is used to retrieve similar entities, and LLM is used to determine whether to merge with a candidate or add as a new entity. A wide range of experiments demonstrate the advanced performance and broad applicability of the AutoMathKG system, including superior reachability query results in MathVD compared to five baselines and robust mathematical reasoning capability in Math LLM.}




\maketitle

\section{Introduction}\label{sec1}

A knowledge graph (KG) aims to construct a knowledge base by extracting a series of triples from unstructured or semi-structured texts \citep{Fensel2020KnowledgeGM}. With the advancement of artificial intelligence (AI) and big data technologies, large-scale KGs have found widespread application in natural language processing (NLP) \citep{zou2020survey}. The mathematical KG, as an important branch, organizes and structures knowledge within the field of mathematics, making it more accessible and usable \citep{zwaneveld2000structuring}. Math KGs can provide a foundational data resource for training more advanced models in mathematical reasoning, promoting the development of AI for mathematics \citep{zhang2023ai}. Since a substantial portion of mathematical knowledge exists in natural language text \citep{szegedy2020promising}, it is of great significance to construct a high-quality and diverse math KG from extensive texts.

Mathematical knowledge can be expressed in two linguistic forms: formal and informal \citep{schoenfeld2012mathematics}. Formal mathematics is written in verifiable formal languages similar to source code, such as Metamath \citep{megill2019metamath} and Lean 4 \citep{moura2021lean}. Informal language, on the other hand, is a blend of symbolic and natural language used by humans. Despite the significant advancements in formal mathematics driven by NLP technologies \citep{wu2021lime}, it does not directly address the informal aspects of mathematics \citep{gowers2010princeton}, failing to promote human understanding of mathematics as effectively as natural mathematical language. Therefore, this paper aims to investigate the construction of a mathematical KG written in natural language.

Existing mathematical KGs in natural language can be mainly categorized as problem-solving \citep{zhao2019mathgraph} and concept-integrating \citep{welleck2021naturalproofs,wang2022math,horowitz2023mathgloss}. While popular, there are two major limitations. First, they are constrained by the completeness of the current corpus, often discarding or manually supplementing missing knowledge, without fully exploring the inherent logic of mathematical knowledge. Second, they are typically static in structure and fail to automatically integrate knowledge from different sources, requiring substantial time and human resources for updates. Addressing these drawbacks, we extend the research on math KGs written in natural language and propose a novel method to construct automatically updatable KG based on large language models (LLMs).

This paper proposes AutoMathKG, a high-quality, wide-coverage, and multi-dimensional KG for the field of mathematics, written in natural language and capable of automatic update. AutoMathKG covers multiple mathematical domains and levels through integrating diverse math sources, including web data from ProofWiki, book data from math textbooks, paper data from arXiv, and problem data from TheoremQA annotated by GPT-4 in MathInstruct \citep{yue2023mammoth}. Innovatively, AutoMathKG regards mathematics as a vast multi-dimensional network composed of definitions, theorems, and problems. LLM is used to enhance math knowledge and extract potential reference relationships through in-context learning (ICL) \citep{brown2020language}, organizing and storing all information in a structured manner. Furthermore, AutoMathKG updates automatically through two designed mechanisms based on vector database (VD) and LLM, facilitating the expansion for math knowledge.

AutoMathKG is represented as a directed graph, where vertices denote math entities, categorized into Definition (Def), Theorem (Thm), and Problem (Prob), and directed edges denote the reference relationships between entities. Regarding entity vertices, three levels of information are stored in JSON format \citep{pezoa2016foundations} for different knowledge requirements. First, basic information is extracted in preprocessing, identifying entities and their relationships via rule-based matching. Second, advanced information with step-by-step proofs and derivations is augmented by the Llama-2 \citep{touvron2023llama} model, inspired by the formal language Lean 4 while avoiding overly complex strategies and operations. Third, query information is collected for all entities connected to the current entity. Regarding directed edges, Llama-2 is used to discover associations among all entities and identify tactic labels when entities are referenced through ICL, similar to the use of tactics in Lean 4.

Considering that similar entity searches through graph traversal are expensive, we design a VD for AutoMathKG, named MathVD. By vectorizing, entity similarity is transformed into vector similarity, making fuzzy search available for mathematical knowledge. We propose two strategies for embedding entity vectors with SBERT \citep{reimers2019sentence}, which differ in how descriptive statements about entities are constructed, resulting in MathVD1 and MathVD2. AutoMathKG updates automatically through two proposed mechanisms. The first mechanism achieves automatic knowledge completion by building Math LLM, which is capable of solving various math problems to supplement incomplete proofs and solutions for new Theorem and Problem entities. The second mechanism enables automatic knowledge fusion from different sources, utilizing MathVD to search for similar candidate entities and Llama-2 to determine whether to merge with the candidate or add as a new entity.
A wide range of experiments and case studies demonstrate superior reachability query results in MathVD compared to five baselines, as well as strong mathematical reasoning capabilities in Math LLM, highlighting the advanced performance of the AutoMathKG system. To sum up, our main contributions are as follows:
\begin{itemize}
\item We construct AutoMathKG, a high-quality, wide-coverage, and multi-dimensional math KG, which can be automatically updated based on VD and LLMs. 
\item We propose two novel math entity embedding strategies for MathVD construction, enabling fuzzy search for math entities.
\item We propose an automatic KG update method involving two mechanisms. One is for knowledge completion by interacting with a designed Math LLM, and another is for knowledge fusion by MathVD and Llama-2. 
\item Experiments demonstrate the advanced performance and broad applicability of the AutoMathKG system, including superior reachability query results in MathVD compared to five baselines and robust mathematical reasoning capability in Math LLM.
\end{itemize}

The remainder of this paper is organized as follows. Section \ref{sec2} introduces related work. Section \ref{sec3} provides an overview of the AutoMathKG system. Section \ref{sec4} describes the KG construction. Section \ref{sec5} details the VD construction. Section \ref{sec6} explains the automatic updates method. Section \ref{sec7} presents the experiments and evaluations. Section \ref{sec8} conducts a further discussion. Section \ref{sec9} concludes our work.

\section{Related work}\label{sec2}

\subsection{Mathematical data resource}

Regarding math KGs, there are two categories: one for solving problems and the other for integrating concepts. MathGraph \citep{zhao2019mathgraph} aims at solving high school math problems automatically. Math-KG \citep{wang2022math} integrates math resources including Baidu Baike and Wikipedia through a pipeline approach. NaturalProofs \citep{welleck2021naturalproofs} comprises math statements and their proofs to construct a reference graph with their reference links based on rule matching. MathGloss \citep{horowitz2023mathgloss} is a linked database of undergraduate-level math concepts leveraging online resources.
Regarding math corpora, there are various sources.
AMPS \citep{hendrycks2021measuring} is a synthetic exercise set while lacks challenging math topics. ProofPile \citep{azerbayev2023proofnet} focuses on math theorem proving but limited to formal proofs. OpenWebMath \citep{paster2023openwebmath} is a large-scale corpus composed of web pages. MathPile \citep{wang2023generative} integrates resources from textbooks, web pages, and others. 
Regarding math benchmarks, there are various datasets for different levels of knowledge. Some datasets focus on elementary math problems, such as NumGLUE \citep{mishra2022numglue} and GSM8K-RFT \citep{yuan2023scaling}, while some are more challenging at the middle and high school level, such as Camel-Math \citep{li2023camel} and MATH \citep{hendrycks2021measuring}. Recently, TheoremQA \citep{chen2023theoremqa} addresses university-level problems. MathInstruct \citep{yue2023mammoth} encompasses various problems with reasoning annotations based on the Chain of Thought (CoT) \citep{wei2022chain} and Procedure of Thought (PoT) \citep{chen2022program} approaches. GHOSTS \citep{frieder2024mathematical} aims to assess advanced mathematical abilities of LLMs.

\subsection{Knowledge graph and vector database}

Modern knowledge graphs (KGs) assert facts in the form of \textit{(subject, predicate, object)} triplets, typically stored in a graph database and visualized in graph structures, where subjects and objects are modeled as nodes, and the relationships between them, namely the predicates, are modeled as edges \citep{ji2021survey}. Some popular KGs include DBpedia \citep{auer2007dbpedia}, YAGO \citep{suchanek2007yago}, YAGO2 \citep{hoffart2013yago2}, and Google Knowledge Graph \citep{singhal2012introducing}. KGs enable graph retrieval augmented generation (RAG) and conversational document retrieval \citep{lewis2020retrieval}. 
An important task is to search for similar entities given an input entity. Current methods for semantic similarity retrieval in KG, such as ontology matching and pattern matching, face challenges with the expansion of KG scale \citep{zheng2016semantic}.
On the other hand, as word embedding techniques emerge, there are many methods to map words into continuous vector space, such as Word2vec \citep{mikolov2013efficient}, GloVe \citep{pennington2014glove}, BERT \citep{devlin2018bert}, and SBERT \citep{reimers2019sentence}. Since closely connected embeddings are semantically related in vector space, the similarity between entity vectors in the vector database can be utilized to retrieve similar entities, and various popular KG embedding models have been proposed, such as TransE \citep{bordes2013translating}, KG2E \citep{he2015learning}, HoLE \citep{nickel2016holographic}, R-GCN \citep{schlichtkrull2018modeling}, and BoxE \citep{abboud2020boxe}.

\subsection{In-context learning with LLMs}

Large Language Models (LLMs) typically refer to transformer language models containing billions (or more) of parameters, trained on massive text data, such as GPT-4 \citep{achiam2023gpt}, PaLM \citep{chowdhery2023palm}, Galactica \citep{taylor2022galactica}, Llama-2 \citep{touvron2023llama}, and Gemma \citep{team2024gemma}. With the increase in model scale and corpus size, LLMs demonstrate powerful capabilities in understanding natural language and exhibit the ability of in-context learning (ICL), where LLMs learn to predict based solely on the in context added a few examples \citep{brown2020language}. ICL has emerged as a new paradigm in NLP. First, some examples are required to form demonstration contexts, often written in natural language templates. Then, a query question is concatenated with a demonstration context to form a prompt. Finally, LLMs are prompted for prediction.
LLMs can perform a variety of complex tasks through ICL, exhibiting strong few-shot performance across many NLP tasks, such as question answering, information extraction, and mathematical reasoning. \citet{olsson2022context} analyzed the mechanisms underlying this behavior. \citet{saunshi2020mathematical} argued that, by conditioning on prompts, the task of predicting the next word was close to linearly separable. \citet{liu2021makes} investigated the impact of the number of provided examples. \citet{min2022rethinking} demonstrated the effectiveness of structuring prompts in an input-label pairing format. \citet{lampinen2022can} suggested that incorporating explanatory task instructions within the context can further enhance performance.

\section{AutoMathKG system overview}\label{sec3} 

\begin{figure}[h]
\centering
\includegraphics[width=0.7\textwidth]{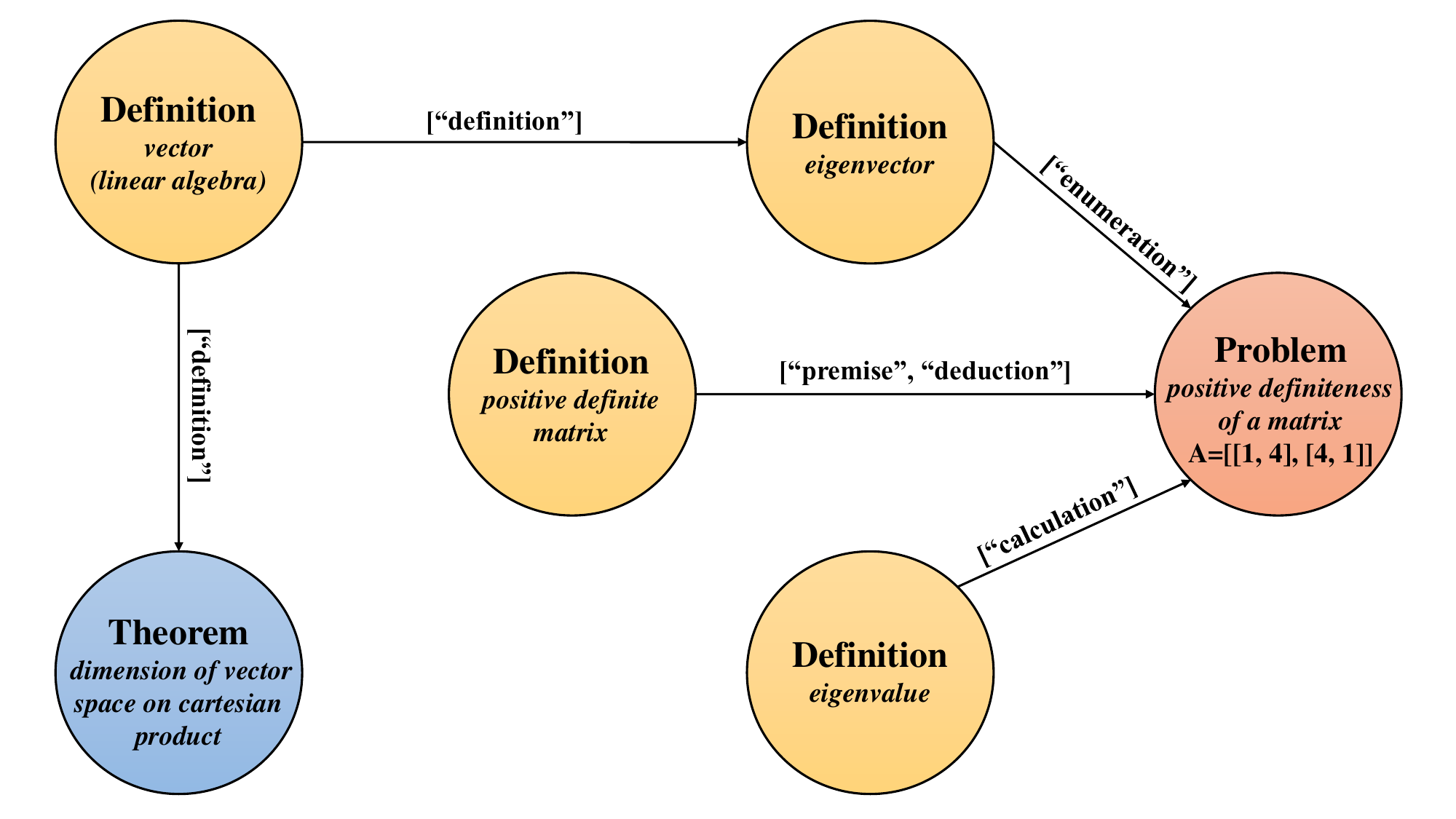}
\caption{Examples of entities and reference edges with tactics in AutoMathKG.}\label{fig1}
\end{figure}

In order to fully explore mathematical knowledge from text and store it in a structured manner, we proposed AutoMathKG, an automatically updatable KG in the field of mathematics written in natural math language. We collect corpora from four sources: ProofWiki, textbooks, arXiv papers, and the TheoremQA dataset, covering a wide range of math domains and levels, and use LLMs to enhance knowledge. AutoMathKG is designed as a directed graph, where vertices represent math entities and directed edges represent reference relationships between entities. Math entities are classified into three types: Definition (Def), Theorem (Thm), and Problem (Prob). Figure \ref{fig1} illustrates examples of some entities and their relationships in AutoMathKG. To achieve our design, there are three main problems: problem 1 is how to extract math entities and relationships from natural language texts, problem 2 is how to effectively augment and store key information about mathematical entities, and problem 3 is how to automatically update entities and their relationships.

As for problem 1, regarding Definition and Theorem entities, we preprocess each collected corpus using the same rule-based matching approach as NaturalProofs dataset, extracting basic information about mathematical definitions, theorems, and theorem proofs, along with extracting reference relationships between them. Regarding Problem entities, since NaturalProofs does not involve math problems, we utilize the problems and their solutions from TheoremQA annotated by GPT-4 in MathInstruct. To associate Problem entities with Definition and Theorem entities, LLM is used to discover potential references to definitions and theorems in Problem entities through ICL. Thus, reference relationships among all kinds of entities are comprehensively extracted. Furthermore, when entities are referenced, we innovatively employ LLM to identify the tactic labels and store them in the directed edges, such as ``premise", ``assumption", ``lemma" and so on. These tactic labels for math entities in natural language form clarify how references drive theorem proofs or problem solutions, similar to the use of tactics in the formal language Lean 4 to assist in constructing theorem proofs.

\begin{figure}[t]
\centering
\includegraphics[width=0.7\textwidth]{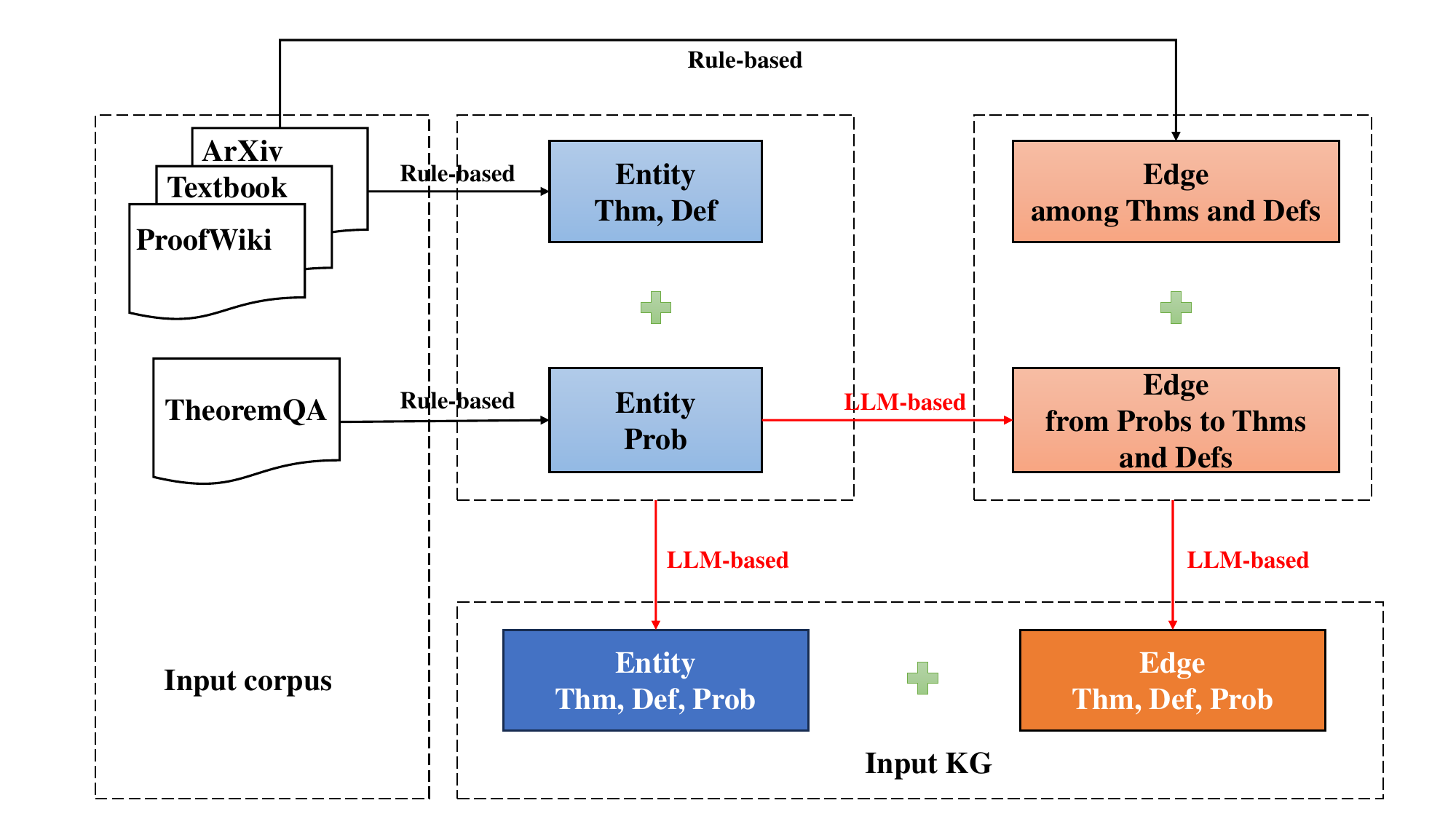}
\caption{Model architecture of AutoMathKG construction for building Input KG from input text.}\label{fig2}
\end{figure}

As for problem 2, we store three levels of information for each entity in order to meet the knowledge requirements in different scenarios. Firstly, basic information is obtained directly from the corpus during entity extraction preprocessing, including entity type, label, title, contents, and source. Secondly, advanced information provides step-by-step proofs and derivations, using LLM to divide theorem contents and proofs into logical segments, inspired by Lean 4 while eliminating overly complex strategies and operations. Lastly, query information contains all incoming and outgoing entities related to the current entity. The three levels of information capture the intrinsic logic of mathematical entities in both content statements and reference citations. All entity information is stored in JSON format, where each key-value pair represents an information attribute of the entity. Figure \ref{fig2} illustrates the flowchart for extracting and storing information from the input corpus to construct the Input KG.

\begin{figure}[t]
\centering
\includegraphics[width=0.7\textwidth]{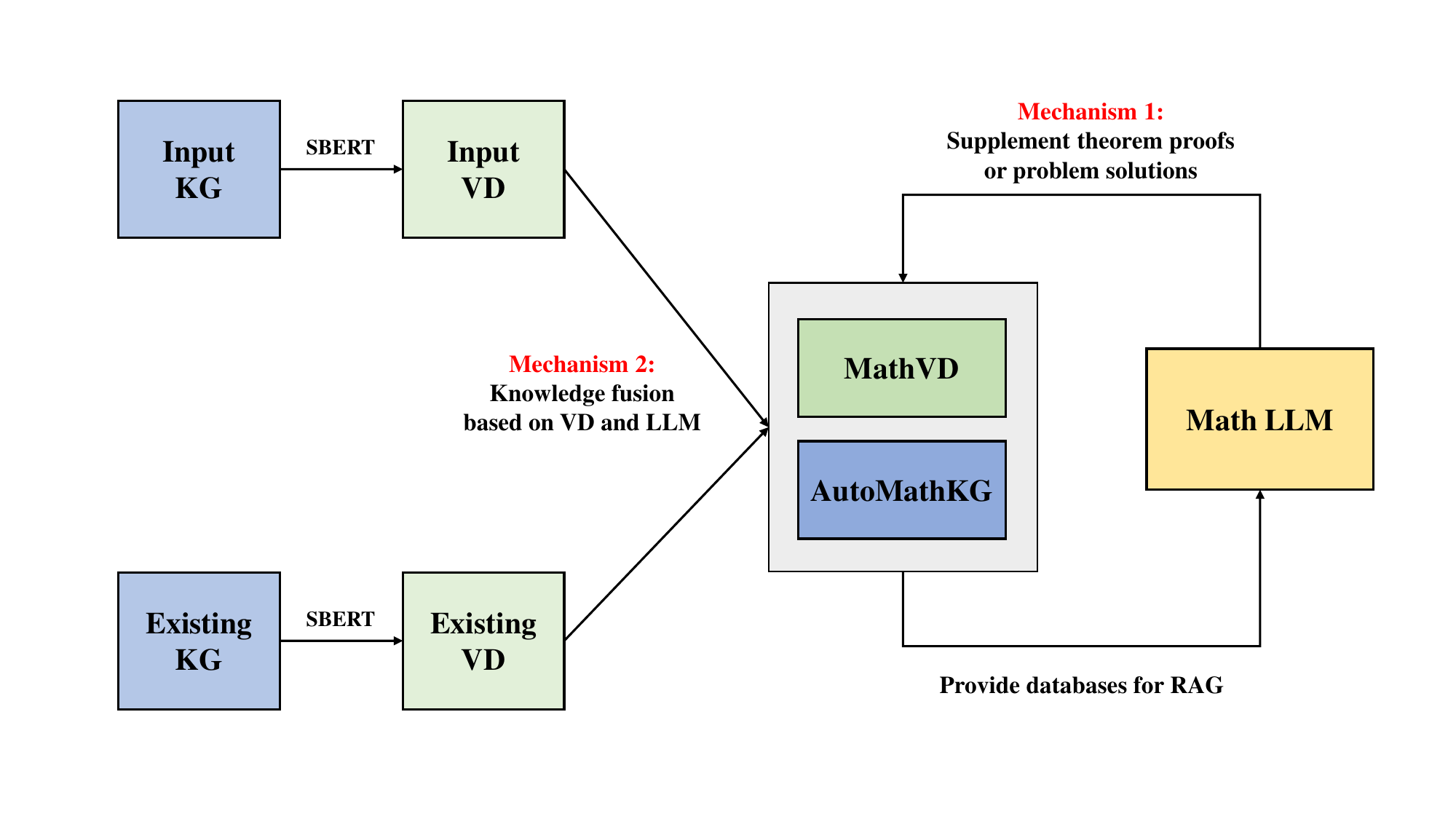}
\caption{Model architecture of AutoMathKG automatic updates for knowledge completion by Math LLM and knowledge fusion by VD and LLM between Input KG and Existing KG.}\label{fig3}
\end{figure}

As for problem 3, we design two mechanisms to achieve automatic updates of our math KG. The first mechanism is to supplement the incomplete knowledge of new entities. We build a specialized mathematical LLM called Math LLM, capable of solving various mathematical problems, to complete missing proofs or solutions for new entities. The second mechanism is to add entities and relationships from new texts in different corpus sources to the Existing KG without duplication or omission. We utilize VD and LLM to update the new entities. First, an Input KG is constructed by extracting entities and relationships from the new input text. Then, Input VD and Existing VD are built from their respective KGs, allowing similar entity candidates for each input entity vector to be retrieved. Finally, through ICL, LLM is employed to determine whether to merge the input entity with a similar candidate or add it as a new entity. Figure \ref{fig3} depicts the flowchart for the automatic updates of our math KG.

\section{Method of AutoMathKG construction}\label{sec4}
\subsection{Corpus collection}\label{sec4-1}

To construct AutoMathKG, we collected diverse data from four sources, including web data from ProofWiki, problem data from TheoremQA, book data from high-quality mathematics textbooks, and paper data from arXiv, covering a wide range of math subjects. Table \ref{t1} presents the corpus statistics for AutoMathKG entities, including Definition (Def), Theorem (Thm), and Problem (Prob) entities, as well as other entities with uncertain types.

\begin{table}[h]
\caption{Statistics of AutoMathKG corpus sources.} \label{t1}
\begin{tabular*}{\textwidth}{@{\extracolsep\fill}llllll@{\extracolsep\fill}}
\hline
\multirow{2}{*}{\textbf{Courpus}}  &\multirow{2}{*}{\textbf{Sample}}       & \multicolumn{4}{c}{\textbf{Entity}}                      \\
 &       & \textbf{Def} & \textbf{Thm} & \textbf{Prob} & \textbf{Other} \\
\hline
ProofWiki         &9496      & 4743         & 1811         & 0             & 2942           \\
Textbook          &1605      & 361          & 1209         & 0             & 35             \\
ArXiv             &538      & 134          & 399          & 0             & 5              \\
TheoremQA         &1749      & 0            & 0            & 1084          & 665            \\
\hline
\textbf{Total}  &13388     & 5238         & 3419         & 1084          & 3647           \\
\hline
\end{tabular*}
\end{table}

\subsubsection{ProofWiki}
ProofWiki\footnote{\url{https://proofwiki.org/}.} is an online compendium of mathematical proofs, containing mathematical definitions, theorems, and their proofs across various domains, contributing significantly to the advancement of formal proof fields. By manually designing rules to filter pages, information such as type, title, contents, and references for each page can be obtained, similar to the approach used in NaturalProofs \citep{welleck2021naturalproofs} and ProofPile \citep{azerbayev2023proofnet}. This paper utilized a portion of ProofWiki data processed by NaturalProofs as the initial source of Definition and Theorem entity data for AutoMathKG.

\subsubsection{TheoremQA}
TheoremQA \citep{chen2023theoremqa} is a theorem-driven QA dataset built upon university-level theorems sourced from various subfields such as algebra, number theory, graph theory, and information theory. \citet{chen2023theoremqa} searched for problems related to these theorems from different sources and standardized the problems to ensure uniform solution formats. \citet{yue2023mammoth} supplemented TheoremQA with reasoning annotations using GPT-4 in MathInstruct, providing solutions through both CoT and PoT methods. This paper utilized TheoremQA data from the MathInstruct dataset with annotated rationales as the initial source of Problem entity data for AutoMathKG.

\subsubsection{Textbook}
Mathematical textbooks contain rich and rigorous math definitions, theorems, proofs, and explicit reference relationships. We conducted extensive manual searches on the internet, specifically targeting open-source mathematical textbooks available on freely accessible websites. For each textbook, we downloaded the \LaTeX\ source code and extracted statements and proofs based on environment names. In total, we processed eight textbooks covering a variety of subjects, including real analysis, algebra, differential equations, elementary number theory, calculus, geometry, and others (see Table \ref{tA1} in Appendix \ref{secA1} for more details). Among these, two textbooks were selected from NaturalProofs, while the remaining six were processed through our work. The collected textbooks were used as a data source for updating AutoMathKG.

\subsubsection{ArXiv}
ArXiv\footnote{\url{https://arxiv.org/}.} is a free distribution service and an open-access archive for millions of scientific papers, providing valuable training data for many powerful language models. We collected 20 mathematical papers from arXiv, downloading the \LaTeX\ source code for each paper and extracting statements and proofs based on environment names. These papers cover various subfields such as algebraic geometry, algebraic topology, differential geometry, statistics, and probability theory (see Table \ref{tA2} in Appendix \ref{secA1} for more details). The collected mathematical papers from arXiv were used as a data source for updating AutoMathKG.

\subsection{AutoMathKG structure}
AutoMathKG is represented as a directed graph $G=\{V,E\}$, where \( V \) is the set of vertices corresponding to mathematical entities, and \( E \) is the set of edges representing the relationships between mathematical entities. Below, we elaborate on the design and storage for vertices and edges.

\subsubsection{Entity vertices}

There are three types of entities in AutoMathKG: Definition entities, Theorem entities, and Problem entities, corresponding to three types of vertices in the directed graph \( G \). 

Definition entities represent concise and comprehensive statements of mathematical concepts. Mathematical definitions serve as the foundation for mathematical theorems and their proofs, as well as the prerequisites for the formulation and solution of mathematical problems. Some complex mathematical definitions may refer to other fundamental definitions. For instance, defining a right triangle involves referencing the definition of a triangle. 

Theorem entities represent mathematical statements that have been logically proven to be true, such as the \textit{Pythagoras’s Theorem}. Since there is a variety of terms indicating the roles of mathematical statements in specific topics, we consider Theorem entities to include theorems, propositions, corollaries, and lemmas. Proving theorems is a central activity in mathematics, and a theorem may have multiple proofs from different perspectives. Hence, in AutoMathKG, Theorem entities include not only the statement of the theorem but also the corresponding proof process.

Problem entities represent mathematical problems that can be formulated, analyzed, and potentially solved using mathematical methods. These problems may range from real world problems, such as calculating the orbits of planets in the solar system, to abstract problems, such as determining the order of a prime group. In the statement and solution of problems, references to other mathematical definitions or theorems are often made, which directly demonstrates the application of mathematical definitions and theorems. Hence, in AutoMathKG, Problem entities include not only the statement of the problem but also the corresponding solution process.

\subsubsection{Entity storage}

To effectively store key information about mathematical entities, every entity is stored in JSON format, where each key-value pair represents an attribute of the entity. The following entity attributes are selected to store: ``id”, ``type”, ``label”, ``title”, ``field”, ``contents”, ``bodylist”, ``refs”, ``references\_tactics”, ``source”, ``proofs”, ``solutions”, ``in\_refs”, ``in\_ref\_ids”, ``out\_refs” and ``out\_ref\_ids”. The entity storage structure is depicted in Fig. \ref{fig4}, and detailed information on all attributes can be found in Table \ref{tA3} of Appendix \ref{secA2}. 

Specifically, ``contents” represents the entity content, presented as a list of sequence strings, with each sequence being a mixture of text and \LaTeX\ language. ``Bodylist” denotes content segmentation, structured as a nested list of dictionaries, where each dictionary provides a description of the segmented content and its corresponding action label, stored in the keys of “description” and “action”, respectively. ``References\_tactics” signifies referenced entities and tactic labels, presented as a dictionary, with each key representing a referenced entity and its value indicating the associated tactic label. We assign the same labels to the action and tactic attributes, which are described in Section \ref{4.2.3}. Table \ref{t2} illustrates examples of content segmentation and reference tactics for Definition, Theorem, and Problem entities. ``In\_refs” and ``out\_refs” respectively store information about all incoming and outgoing edge entities of the current entity. Additionally, Theorem entities include ``proofs”, and Problem entities include ``solutions”. Each theorem proof and problem solution contains ``contents”, ``bodylist” (only in theorem proofs), ``refs” and ``references\_tactics”, following the same format as above. An instance of a Definition entity is shown in Fig. \ref{fig9} of Appendix \ref{secA5}.

\begin{table}[]
\caption{Samples of different types of entities storage.}\label{t2}
\begin{tabular*}{\textwidth}{@{\extracolsep\fill}lc@{\extracolsep\fill}}
\hline
\multicolumn{2}{l}{\textbf{Definition: Symmetric Group}} \\ \hline
Contents                   & \makecell[l]{Let S be a set. Let $\Gamma(S)$ denote the set of permutations on S. \\ Let $(\Gamma ,\circ)$ be the algebraic structure such that $\circ$ denotes the \\composition of mappings. Then $(\Gamma ,\circ)$ is called the symmetric \\group on S. If S has n elements, then $(\Gamma ,\circ)$ is often denoted $S_n$. } \\  
\hline
Bodylist                             & \makecell[l]{[\{``description": ``Let S be a set.”, ``action": ``premise"\},\\
\{``description": ``Let $\Gamma(S)$ denote the set of permutations on S. \\ Let $(\Gamma ,\circ)$ be the algebraic structure such that $\circ$ denotes the \\composition of mappings. Then $(\Gamma ,\circ)$ is called the symmetric \\group on S. If S has n elements, then $(\Gamma ,\circ)$ is often denoted $S_n$.”, \\``action": ``definition"\}]}
                 \\
\hline
References\_tactics                  &  \makecell[l]{\{``Definition:Composition of Mappings": ``definition",\\
``Definition:Algebraic Structure": ``definition",\\
``Definition:Element": ``definition",\\
``Definition:Permutation": ``definition",\\
``Definition:Set": ``premise"\}
}                 \\
\hline\hline
\multicolumn{2}{l}{\textbf{Theorem: Center of Symmetric Group is Trivial}} \\ 
\hline
Contents                   & \makecell[l]{Let $n\in \mathbb{N}$ be a natural number. Let $S_n$ denote the symmetric group \\ of order n. Let $n\ge 3$. Then the center $Z(S_n)$ of $S_n$ is trivial. } \\  
\hline
Bodylist                             & \makecell[l]{[\{``description": ``Let $n\in \mathbb{N}$ be a natural number.”, ``action": ``premise"\},\\ \{``description": `` Let $S_n$ denote the symmetric group \\ of order n. Let $n\ge 3$.”, ``action": ``assumption"\},\\
\{``description": ``Let $n\ge 3$. Then the center $Z(S_n)$ of $S_n$ is trivial.”, \\``action": ``conclusion"\}]}
                 \\
\hline
References\_tactics                  &  \makecell[l]{\{``Definition:Order of Structure": ``premise",\\
``Definition:Center (Abstract Algebra)/Group": ``premise",\\
``Definition:Natural Numbers": ``premise",\\
``Definition:Trivial Group": ``premise",\\
``Definition:Symmetric Group": ``premise"
\}
}                 \\
\hline\hline
\multicolumn{2}{l}{\textbf{Problem: Positive Definiteness of a Matrix}} \\ \hline
Contents                   & \makecell[l]{Consider the matrix of $A=[[1, 4], [4, 1]]$, is this a positive definite matrix?} \\  
\hline
References\_tactics                             & \makecell[l]{\{``definition:positive definite matrix": ``deduction"\}}
                 \\
\hline
\end{tabular*}
\end{table}

\begin{figure}[h]
\centering
\includegraphics[width=0.7\textwidth]{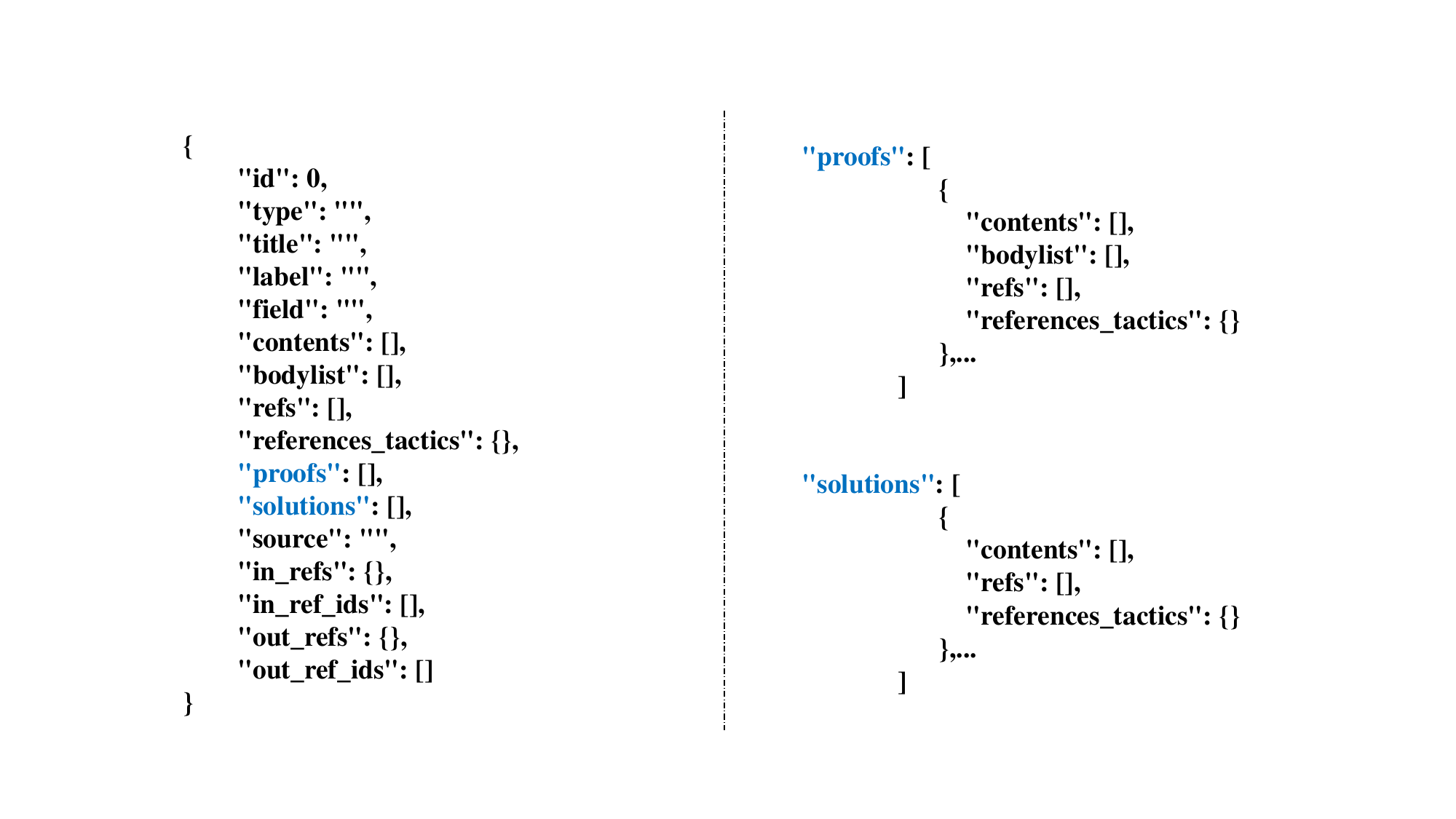}
\caption{ The schema of entity storage in JSON format.}\label{fig4}
\end{figure}

\subsubsection{Entity relation edges}\label{4.2.3}

The edges in AutoMathKG are directed edges, representing reference links. An edge from entity vertex $v_i \in V$ to entity vertex $v_j \in V$, denoted as $e_{i\to j}  \in E$, implies that entity $v_j$ references entity $v_i$ in its content statement, proof process, or solution process. Edges can point to any type of entity, but they always originate from a Definition or Theorem entity. These directed edges form logical or inferential chains among mathematical knowledge, demonstrating the dependency relationships and deduction processes between different mathematical entities.

\begin{figure}[h]
\centering
\includegraphics[width=0.7\textwidth]{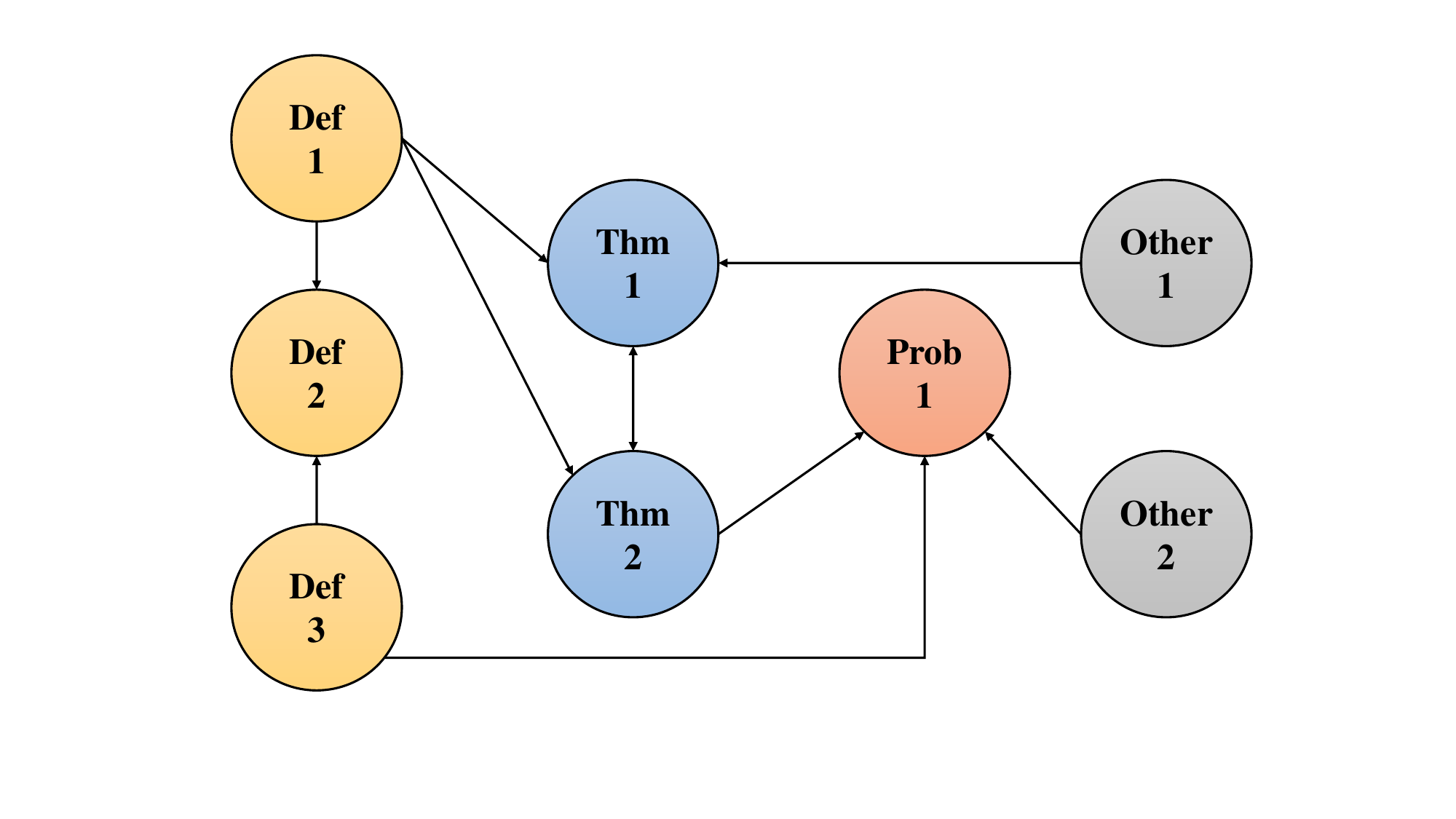}
\caption{The diagram of all possible relationships between different types of entities.}\label{fig5}
\end{figure}

All references between entities form a directed graph, which may contain cycles. For example, \textit{Pythagoras’s Theorem} and \textit{Sum of Squares of Sine and Cosine} are mutually referenced in their proofs. Figure \ref{fig5} illustrates all possible relationships between different types of entities. Additionally, to clarify the tactic role played by entities when referenced, we design nine tactic labels: premise, assumption, lemma, proposition, corollary, calculation, enumeration, definition, and conclusion. Each edge includes tactic label information, stored in the ``in\_refs" and ``out\_refs" fields of the entity JSON structure.

\subsection{Information extraction and augmentation}\label{sec4.4}

This subsection will detail the instruction on how to extract and augment information to construct an Input KG according to the input text from a technical perspective.

\paragraph{Step 1: Rule-based extraction}  

For each \LaTeX\ source code of the input text, rule-based matching method \citep{welleck2021naturalproofs}, depending on \LaTeX\ environment names, is conducted to extract basic information of different types of entities. Specifically, it searches for predefined environment names and parses the content contained in these environments into basic information corresponding to the types of entities. For instance, the content surrounded by \verb|\begin{theorem}| and \verb|\end{theorem}| in the input text will be parsed as the content of Theorem entities in AutoMathKG. After rule-based matching, most of the following attribute information of entities can be obtained for the current input text: ``id", ``type", ``label", ``title", ``contents", ``refs", ``source", ``proofs" and ``solutions".

\paragraph{Step 2: LLM-based augmentation}

After rule-based matching, each entity undergoes content and relationship augmentation by LLM using ICL to complete the remaining attribute information. Specifically, for attributes such as ``title", ``field", ``bodylist", ``references\_tactics" and ``refs", prompt templates are designed to ask LLM for the information corresponding to each attribute. Considering the different requirements of attribute prompt templates for different entity types, we design a total of 12 prompt templates (see Table \ref{tA4} in Appendix \ref{secA3} for more details). Finally, the fully augmented information of each entity is stored in JSON format, resulting in an Input KG generated from the current input text.

\section{Method of MathVD construction}\label{sec5}

Given that similarity searches through graph traversal become increasingly expensive as the scale of the KG expands, we build a VD for AutoMathKG, denoted as MathVD. Through vectorizing the query content and the KG, the similarity between entities is measured through the cosine similarity of their respective vectors, enabling fuzzy search for math knowledge. SBERT \citep{reimers2019sentence} is employed to embed the descriptions of math entities, generating 384-dimensional entity vectors. Two strategies for embedding entity vectors are designed, differing in how descriptive statements about entities are constructed, resulting in two VDs: MathVD1 and MathVD2. 

\subsection{MathVD1}\label{subsec-MathVD1}\label{sec5.1}

For each entity, MathVD1 uses SBERT to embed a long text containing all key information about the entity to generate its vector. First, it is necessary to determine what information is vital to represent math entities. For any entity node $v$, five important fields containing entity features from its JSON format storage are selected, namely ``title", ``field", ``contents", ``in\_refs", and ``out\_refs". The values stored in each field are obtained and presented as descriptive sentences $S_1{(v)},\cdots,S_5{(v)}$, given by 
\begin{align}
S_1{(v)} &= ``title: \; v[title]", \nonumber \\
S_2{(v)} &= ``field: \; v[field]", \nonumber \\
S_3{(v)} &= ``content: \; v[contents]", \nonumber \\
S_4{(v)} &= ``in \; references: \; v[in\_refs]", \nonumber \\
S_5{(v)} &= ``out \; references: \; v[out\_refs]", \label{eq1}
\end{align}
where $v[key]$ represents the value of the $key$ field for the entity $v$ stored in KG. Subsequently, MathVD1 concatenates each descriptive sentence sequentially to form a long text $S{(v)}$  containing all key information about the entity, given by
\begin{align}
S{(v)} = S_1{(v)}+S_2{(v)}+S_3{(v)}+S_4{(v)}+S_5{(v)}. \label{eq2}
\end{align}
Finally, the long sentence $S{(v)}$ is embedded by SBERT, obtaining the vector representation $e{(v)} \in \mathbb{R} ^{384}$.

MathVD1 considers $e(v)$ as the vector embedding of the entity $v$. Table \ref{t3} illustrates an instance of similarity retrieval for the target entity in MathVD1, where cosine similarity is utilized as the similarity metric, indicating significant relevance of the retrieved content to the target entity.

\begin{table}[h]
\caption{The result of a similar entity retrieval example in MathVD1.}\label{t3}
\begin{tabular*}{\textwidth}{@{\extracolsep\fill}cll@{\extracolsep\fill}}
\hline
\multicolumn{3}{l}{\textbf{Target entity:} Set union is associative: $A\cup(B\cup C)=(A\cup B)\cup C$}                             \\ \hline
\multicolumn{3}{l}{\textbf{Retrieval results in MathVD1}}                             \\ \hline
\multicolumn{1}{c}{\textbf{Rank}} & \textbf{Similar entity} & \multicolumn{1}{c}{\textbf{Score}}\\ 
\hline
1      & Theorem: Set union is commutative:
$S\cup T=T\cup S$ & 0.8977  \\ 
2      & Theorem: Symmetric difference is associative:
$R\triangle (S\triangle T)=(R\triangle S)\triangle T$	
               & 0.8300  \\ 
3      & Theorem: Set intersection is associative: $A\cap(B\cap C)=(A\cap B)\cap C$
               & 0.8217  \\ 
\hline            
\end{tabular*}
\end{table}

\subsection{MathVD2}
For each entity, MathVD2 uses SBERT to separately embed each key information description of the entity, then weights and sums these vectors based on their respective importance to generate the entity vector. First, we need to determine the descriptive sentence of key information for each entity. For any entity node $v$, the same method as described in Section \ref{subsec-MathVD1} is used to obtain key information, resulting in five sentences $S_1{(v)},\cdots,S_5{(v)}$, containing ``title", ``field", ``contents", ``in\_refs" and ``out\_refs" information, according to Equation \ref{eq1}. Subsequently, each sentence is embedded by SBERT, obtaining corresponding vectors $e_i{(v)} \in \mathbb{R} ^{384}$ for each sentence $S_i{(v)},i=1,\cdots,5$. Finally, these vectors are weighted and summed based on the importance of the information. Specifically, let the weight vector $W$ be defined as follows:
\begin{align}
W = (w_1, w_2, w_3, w_4, w_5), \nonumber 
\end{align}
where \( w_j \) represents the weight assigned to the \( j \)-th information, satisfying $ {\textstyle \sum_{j=1}^{5}} w_j = 1$. Then the weighted sum of all vectors is computed by
\begin{align}
\tilde{e}(v)= w_1 \cdot e_1{(v)} + w_2 \cdot  e_2{(v)} + w_3 \cdot  e_3{(v)} + w_4 \cdot  e_4{(v)} + w_5 \cdot  e_5{(v)}.
\end{align}

MathVD2 considers $\tilde{e}(v)$ as the vector embedding of the entity $v$. Since ``contents" can best differentiate between entities, followed by ``title", we set the weight vector as $W = (0.5, 0.3, 0.1, 0.05, 0.05)$ in our work. Similarly, Table \ref{t4} illustrates a similarity retrieval instance for the same target entity in MathVD2, also indicating significant relevance to the target entity. Comparing Tables \ref{t3} and \ref{t4}, we can see that both MathVD1 and MathVD2 are effective in retrieving similar entities, although there are slight differences in the retrieval results. 
Besides, some similar math Problem entities with the same mathematical essence exhibit high cosine similarity between each other in both MathVDs, shown as Table \ref{t4-1-2}.

\begin{table}[h]
\caption{The result of a similar entity retrieval example in MathVD2.}\label{t4}
\begin{tabular*}{\textwidth}{@{\extracolsep\fill}cll@{\extracolsep\fill}}
\hline
\multicolumn{3}{l}{\textbf{Target entity:} Set union is associative: $A\cup(B\cup C)=(A\cup B)\cup C$}                             \\ \hline
\multicolumn{3}{l}{\textbf{Retrieval results in MathVD2}}                             \\ \hline
\multicolumn{1}{c}{\textbf{Rank}} & \textbf{Similar entity} & \multicolumn{1}{c}{\textbf{Score}}\\ 
\hline
1      & Theorem: Set union is commutative:
$S\cup T=T\cup S$ & 0.8034  \\ 
2      & Theorem: Set union is idempotent:
$S\cup S=S$	
               & 0.7662  \\ 
3      & Theorem: Set intersection is associative: $A\cap(B\cap C)=(A\cap B)\cap C$
               & 0.7600  \\ 
\hline            
\end{tabular*}
\end{table}

\begin{table*}[h]
\caption{The similarity results of similar Problem entity pairs in MathVD1 and MathVD2.}\label{t4-1-2}
\resizebox{\textwidth}{!}{
\begin{tabular}{llcc}
\hline
\multicolumn{2}{l}{\multirow{2}{*}{\textbf{Similar Problem entity pair}}} & \multicolumn{2}{c}{\textbf{Similarity score}}                                \\
\multicolumn{2}{l}{}                        & \multicolumn{1}{l}{\textbf{MathVD1}} & \textbf{MathVD2}                   \\
\hline
\multirow{2}{*}{\textbf{Pair 1}}     & Entity 1: Consider the matrix of $A=[[1, 4], [4, 1]]$, is this a positive definite matrix?    & \multirow{2}{*}{0.9346} & \multirow{2}{*}{0.9518} \\
                           & Entity 2: Consider the matrix of $A=[[1, -1], [-1, 4]]$, is this a positive definite matrix?    &                                  &                       \\
\hline
\multirow{2}{*}{\textbf{Pair 2}}         & Entity 3: Are the vectors $[1, 2], [2, 3]$, and $[3, 4]$ linearly independent?    & \multirow{2}{*}{0.8291} & \multirow{2}{*}{0.8846} \\
                           & Entity 4: Are the vectors $v_1 = [1,2,3], v_2 = [4,5,6], v_3 = [7,8,9]$ linearly independent?    &                                  &    \\                  
\hline
\end{tabular}
}
\end{table*}

\section{Method of automatic updates}\label{sec6}

In this section, we propose two mechanisms for automatically updating our math KG, in order to fully leverage existing mathematical knowledge and maintain synchronization with ongoing advancements in the field. In the first mechanism, a specialized mathematical LLM for solving various math problems is designed, named Math LLM, to achieve automatic knowledge completion by supplementing incomplete proofs or solutions. In the second mechanism, VDs and LLM are utilized to realize automatic knowledge fusion by merging and updating new entities and their relationships from different sources.

\subsection{Automatic knowledge completion}

\begin{figure}[h]
\centering
\includegraphics[width=0.7\textwidth]{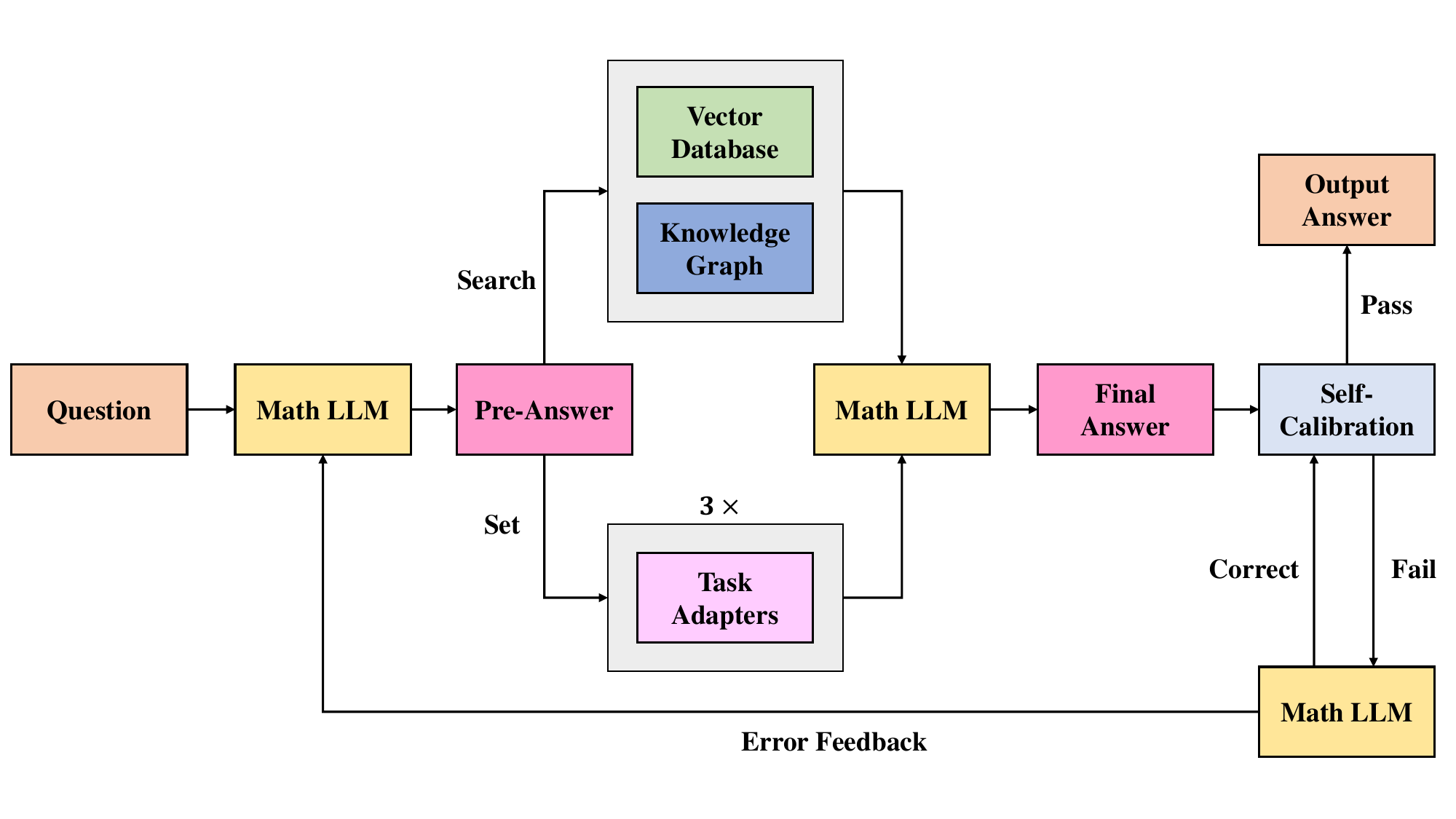}
\caption{Model architecture of Math LLM.}\label{fig6}
\end{figure}

Given that small-scale LLMs with fewer parameters tend to perform less effectively in mathematical reasoning, this paper proposes Math LLM, specifically designed to address various types of mathematical problems through the integration of task adapters, retrieval augmentation, and self-calibration. 
Figure \ref{fig6} illustrates the structure of Math LLM. First, for each mathematical problem input, preliminary information on problem types and knowledge fields is generated. Next, a CoT adapter agent analyzes the information to determine the most appropriate adapter for problem-solving. Meanwhile, relevant knowledge is retrieved from AutoMathKG and MathVD through a two-stage retrieval augmentation. Subsequently, answers are generated using the relevant knowledge as prompts, with the adapter activated. Finally, the self-calibration module reviews and outputs calibrated answers.

\subsubsection{Task adapters}

An adapter is a parameter-efficient fine-tuning strategy that provides plug-and-play functionality, stability, and generalization. It introduces lightweight layers into the model, which are activated only when necessary, allowing the majority of the model parameters to remain unchanged.
Given that, Math LLM employs three types of mathematical expert adapters to address three categories of problems: application, calculation, and proof, utilizing the LORA \citep{hu2021lora}, PoT \citep{chen2022program}, and CoT \citep{wei2022chain} approaches, respectively. 
The application adapter, trained on an application problem dataset, enables the model to abstract real-world problems into mathematical formulations. The calculation adapter, trained on a programming dataset, equips the model with the capability to write Python programs to solve calculation problems, significantly reducing hallucination issues commonly encountered by LLMs in calculations. The proof adapter, trained on a mathematical theorem proof dataset, assists the model in developing a rigorous mathematical reasoning process, thereby strengthening its theoretical framework and logical structure. Finally, a CoT agent is integrated into Math LLM to coordinate these adapters, selecting the most appropriate one for problem-solving (see Appendix \ref{secA4} for more training details).

\subsubsection{Retrieval augmentation}
 
Retrieving knowledge from external databases provides LLMs with verified information, enabling more grounded reasoning and mitigating hallucinations. Thus, Math LLM is designed to interact with AutoMathKG and MathVD for specialized knowledge retrieval. First, Math LLM analyzes the input problem and generates a set of relevant unverified knowledge points, including theorems and definitions. Next, Math LLM extracts relevant knowledge from both databases based on these knowledge points.
Given that exact searches are resource-intensive and fuzzy searches may introduce less precise results, we designed a balanced retrieval augmentation strategy comprising two stages: exact search and fuzzy search. In the first stage, an exact search is performed in AutoMathKG to match Definition and Theorem entities directly to the knowledge points. In the second stage, a fuzzy search is conducted in MathVD to identify entities similar to the knowledge points and those found in the first stage within the vector space. After this, verified theorem and Definition entities are extracted. Similarly, the two-stage retrieval process is used to search for relevant Problem entities across both databases. Finally, all verified auxiliary knowledge entities are packaged and provided to Math LLM, which selects those contributing to the problem-solving process as the final specialized knowledge.

\subsubsection{Self-calibration}

A self-calibration module can effectively reduce hallucinations \citep{zhao2023knowing}, by providing a feedback mechanism for LLMs. Given that, this paper introduces a self-calibration module within Math LLM. Since engaging in trial and error is a common approach for solving unfamiliar and complex problems, our self-calibration module refines the answer incrementally based on CoT. We implement specific principles for calibration, with different types of problems adhering to distinct rules. The model verifies these rules step by step. If the model passes the self-calibration, it outputs the final answer. Otherwise, it identifies the violated principles and attempts to correct the errors. The corrected answer is then fed back into Math LLM for another round of self-calibration. If it still fails, Math LLM generates an error summary, which is combined with the original problem and used as a prompt to initiate another cycle of answer revision until it passes the self-calibration module. 

\subsection{Automatic knowledge fusion}
Considering that mathematical corpora from different sources may have different statements of the same knowledge, we design a mechanism to automatically achieve knowledge fusion. For each new math entity from other sources, we utilize VD to search for similar candidate entities of Existing KG and employ LLM to determine whether to merge with a candidate or add the new entity. Given new input text, the following steps are taken for automatic knowledge fusion.

\paragraph{Step 1: Construction of Input KG from new input}
Information extraction and augmentation are conducted on the new input text as described in Section \ref{sec4.4}, for entity recognition preprocessing and knowledge enhancement, generating Input KG from the current input.

\paragraph{Step 2: Construction of VDs from Input KG and Existing KG}
Entities of Input KG and Existing KG are embedded using SBERT as described in Section \ref{sec5}, constructing two VDs, named Input VD and Existing VD, respectively.

\paragraph{Step 3: Fuzzy search for similar candidate entities in VD}
For each input entity $v_{input}$, a fuzzy search for its similar entities is conducted by comparing the cosine similarity between their entity vectors. Specifically, let $e_{input}$ be the vector embedding of $v_{input}$ in Input VD. Then, for each entity vector in Existing VD, calculate the cosine similarity with $e_{input}$. Finally, sort them in descending order and select the top $n$ entity vectors as the similar candidate entities for the input entity $v_{input}$.

\paragraph{Step 4: Determination of entity consistency by LLM}
After obtaining $n$ similar candidate entities, LLM is used to determine whether $v_{input}$ is consistent with any of the candidates. Specifically, we employ LLM through ICL to determine the consistency (see Table \ref{tA5} in Appendix \ref{secA3} for more details). For each similar candidate entity $v_{i}, i=1,\cdots, n$, LLM is asked to output ``yes" if it is consistent to $v_{input}$; otherwise, LLM outputs ``no".

\paragraph{Step 5: Selection of update approach by LLM}
Finally, we analyze the consistency results and select the appropriate update approach for each input entity. If LLM outputs ``no" for all similar candidate entities, $v_{input}$ will be considered as a new entity and added to the Existing KG. If LLM outputs ``yes" for only one candidate, $v_{input}$ will be merged with it, with reference relationships between $v_{input}$ and other entities in Input KG being added. If LLM outputs ``yes" for multiple candidates, LLM will perform a secondary judgment through ICL on all the candidates with ``yes" output, selecting the most similar candidate entity (see Table \ref{tA5} in Appendix \ref{secA3} for more details). Then, $v_{input}$ will be merged with that candidate, with relationships being added. 

\section{Experiments}\label{sec7}
\subsection{Settings}
As for the construction of AutoMathKG, the Llama-2-7b model\footnote{\url{https://huggingface.co/meta-llama/Llama-2-7b-chat-hf}.} was employed for entity augmentation and all ICL tasks. As for the construction of MathVD, the all-MiniLM-L6-v2 model\footnote{\url{https://huggingface.co/sentence-transformers/all-MiniLM-L6-v2}.} from SBERT \citep{reimers2019sentence} was utilized. When it comes to the automatic update with MathVD, the number of retrieved similar candidate entities was set to $5$, and cosine similarity was employed as the search metric. As for the construction of Math LLM, the gemma-7b-it model\footnote{\url{https://huggingface.co/google/gemma-7b-it}.} was selected as the base model, a lightweight model that demonstrates excellent performance across various tasks \citep{team2024gemma}. 
Initially, the model was fine-tuned on a comprehensive mathematical dataset to develop its basic mathematical problem-solving capabilities. Subsequently, task adapters were trained on different types of datasets. Detailed information regarding the training datasets and hyperparameters is provided in Appendix \ref{secA4}. All experiments were conducted on a computer equipped with four RTX 3090 GPUs and 96GB of VRAM, utilizing Python 3.10.

\subsection{Results}

\begin{figure}[h]
\centering
\includegraphics[width=0.7\textwidth]{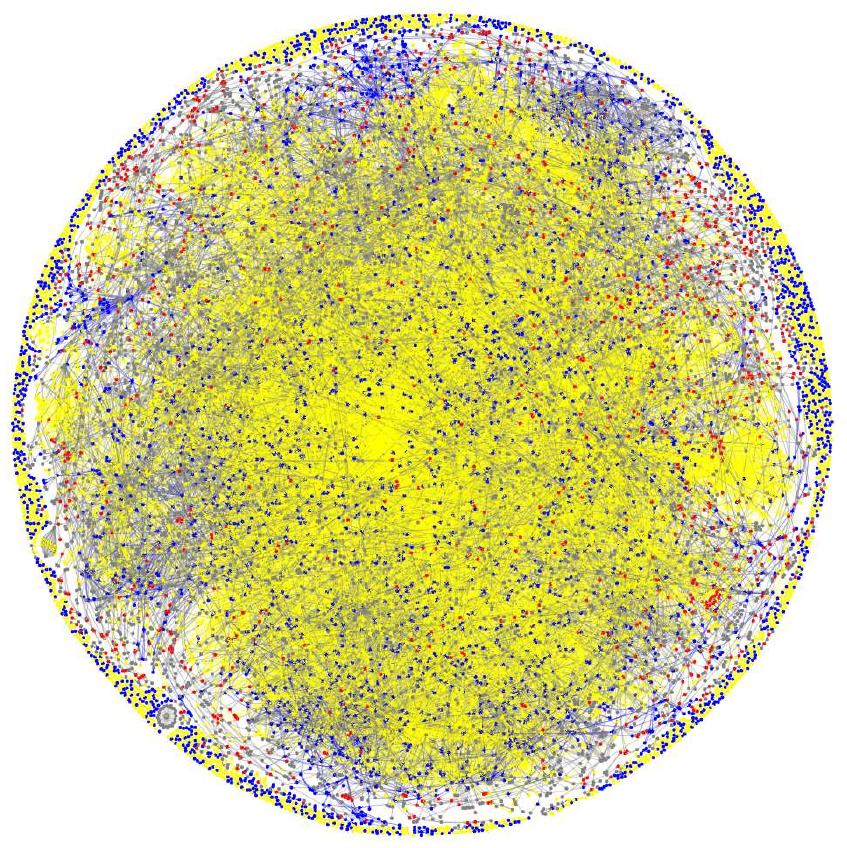}
\caption{The visualization of AutoMathKG, with yellow, blue, and red nodes representing Definition, Theorem, and Problem entities.}\label{fig7}
\end{figure}

\begin{figure}[t]
\centering
\includegraphics[width=0.7\textwidth]{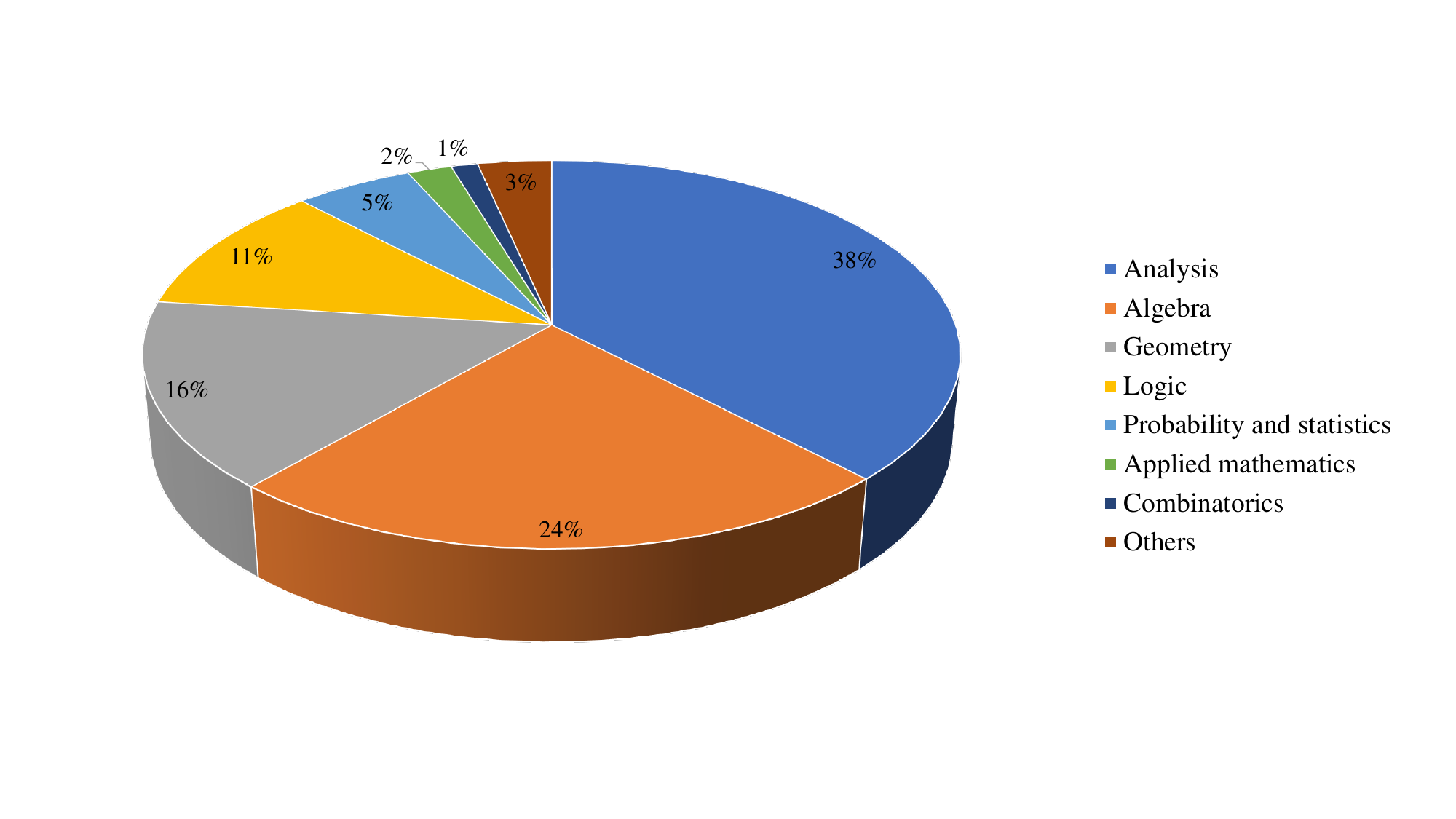}
\caption{Mathematical field distribution in AutoMathKG.}\label{fig10}
\end{figure}

Following the experimental procedure described above, we constructed and updated AutoMathKG. Figure \ref{fig7} illustrates the visualization results generated using the PiVis library, where yellow, blue, and red nodes represent Definition, Theorem, and Problem entities, respectively. An interactive graph for AutoMathKG was also created, with title information displayed on the nodes, allowing users to zoom in and out, as well as move nodes and edges to explore the interconnections between entities\footnote{The interactive graph is available in the related file.}.
AutoMathKG encompasses a broad range of mathematical fields, including Analysis, Algebra, Geometry, Logic, Probability, Statistics, and others, as shown in Fig. \ref{fig10}. Moreover, detailed information regarding the graph structure of AutoMathKG is presented in Table \ref{t5}. The graph is directed and comprises 13,388 entity vertices and 29,459 directed edges. The distribution of head and leaf nodes indicates that many knowledge paths within AutoMathKG start from basic concepts, gradually deepen, and eventually reach specific conclusions. Additionally, the presence of numerous cycles highlights complex relationships and mutual dependence among mathematical knowledge, where simple cycles represent the most fundamental and direct interrelationships.

\begin{table}[h]
\caption{The statistics of graph structure for AutoMathKG.}\label{t5}
\begin{tabular*}{\textwidth}{@{\extracolsep\fill}lccc@{\extracolsep\fill}}
\hline
\textbf{Node} & \textbf{All nodes}     & \textbf{Head node} & \textbf{Leaf node}   \\
\textbf{Number} & 13388                  & 4216                & 3789                  \\
\hline
\textbf{Edge} & \textbf{All edges} & \textbf{Cycle}     & \textbf{Simple cycle} \\
\textbf{Number} & 29459                  & 82153               & 145        \\
\hline
\end{tabular*}
\end{table}

\subsection{Evaluation}

\subsubsection{Reachability query}

To assess the capability of the constructed VDs in capturing structural information from the KG, we conducted reachability query experiments to evaluate whether structurally related entities in the KG could be effectively retrieved based on the similarity between the entity embeddings. Specifically, we considered the $k$-hop reachability query between entities. Two entity nodes $A$ and $B$ are considered $k$-hop reachable if they can be connected by no more than $k$ directed edges in the graph. For any entity vector $e$, we retrieved the top $q$ most similar entities using cosine similarity and counted the number of $k$-hop reachable entities, denoted as $r$. The hit rate for $q$ queries (Hits@$q$) is defined as follows:
\begin{equation}
    Hits@q=\frac{r}{q}.
\end{equation}

We compared MathVD against five baselines, TransE \citep{bordes2013translating}, KG2E \citep{he2015learning}, HoLE \citep{nickel2016holographic}, R-GCN \citep{schlichtkrull2018modeling}, and BoxE \citep{abboud2020boxe}, covering different kinds of KG embedding models. We used PyKEEN\footnote{\url{https://github.com/pykeen/pykeen}.} to implement them, where the vector dimension was set to 384, the same as MathVD. In the experiments, we randomly sampled 100 entities based on the category proportions, including 50 Definition entities, 30 Theorem entities, and 20 Problem entities. Main results of Hits@$q$ for 5-hop reachability query using baselines and our models are shown in Table \ref{t66}. The best results are highlighted in bold. 
\begin{table}[h]
\caption{The results of Hits@$q$ for 5-hop reachability query with $q$ = 1, 5, 10, and 15.}\label{t66}
\begin{tabular*}{\textwidth}{@{\extracolsep\fill}lcccc@{\extracolsep\fill}}
\hline  
\textbf{Model}   & \textbf{Hits@1}                & \textbf{Hits@5}                         & \textbf{Hits@10}                        & \textbf{Hits@15}                        \\
\hline  
TransE  & \textbf{0.9610}               & 0.7766                                 & 0.7182                                 & 0.6797                                 \\
KG2E   & 0.7403                        & 0.7325                                 & 0.7156                                 & 0.7091                                 \\
HoLE    & 0.8442                        & 0.7688                                 & 0.7312                                 & 0.7022                                 \\
R-GCN   & 0.7273                        & 0.7065                                 & 0.6948                                 & 0.6840                                 \\
BoxE    & 0.9351                        & 0.8338                                 & 0.7610                                 & 0.7247                                 \\
\hline  
MathVD1     & 0.8831                        & 0.8364                                 &  \textbf{0.8182} &  \textbf{0.7861} \\
MathVD2     & 0.8974 &  \textbf{0.8385} & 0.8013                                 & 0.7786    \\       \hline                     
\end{tabular*}
\end{table}

\begin{figure*}[h]
\centering  
\subfigure[MathVD1]{   
\setcounter{subfigure}{0}
\begin{subfigure}
\centering    
\includegraphics[width=0.46\linewidth]{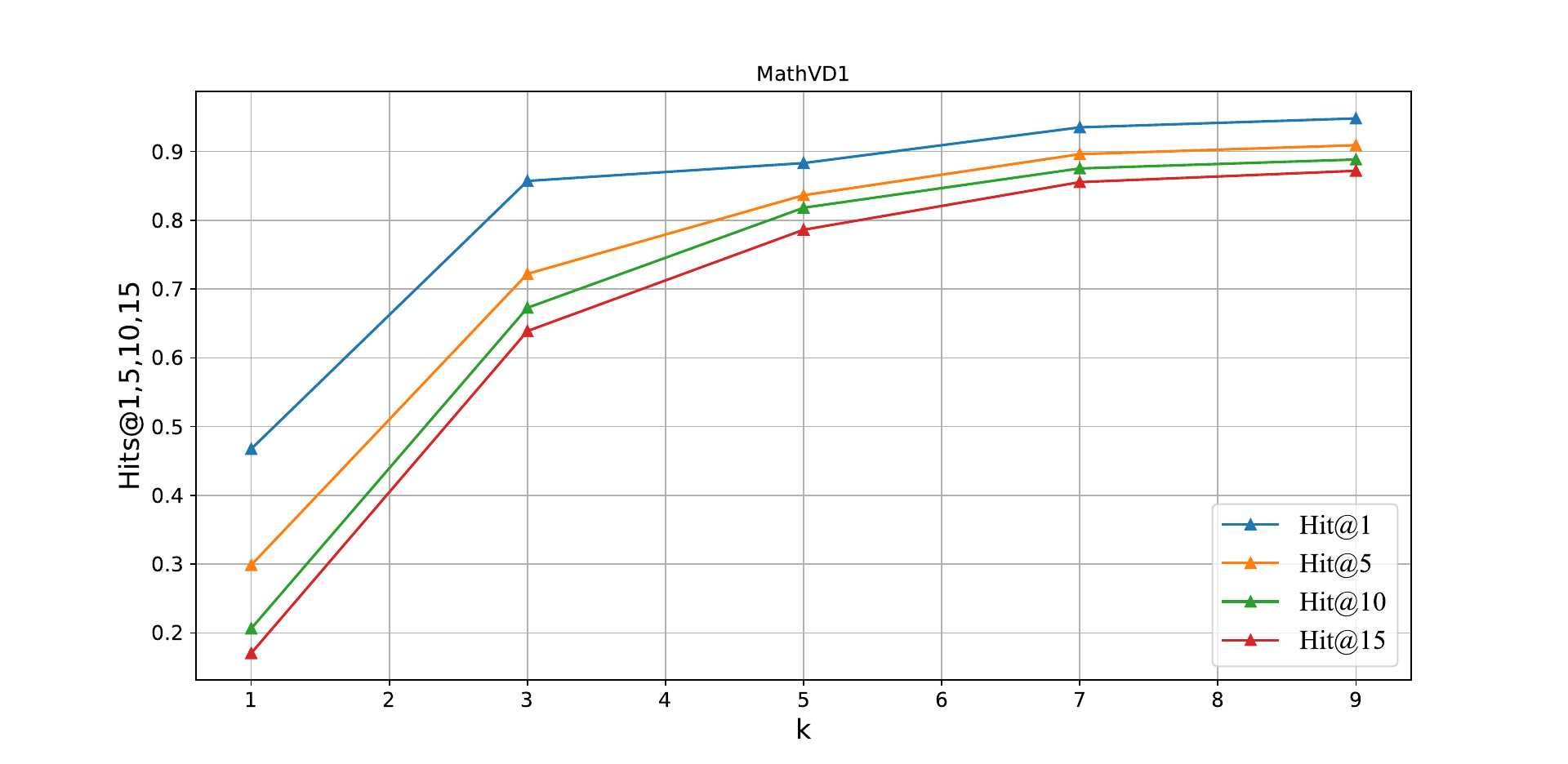}
\label{fig_MathVD1}
\end{subfigure}
}
\subfigure[MathVD2]{ 
\setcounter{subfigure}{1}
\begin{subfigure}
\centering    
\includegraphics[width=0.46\linewidth]{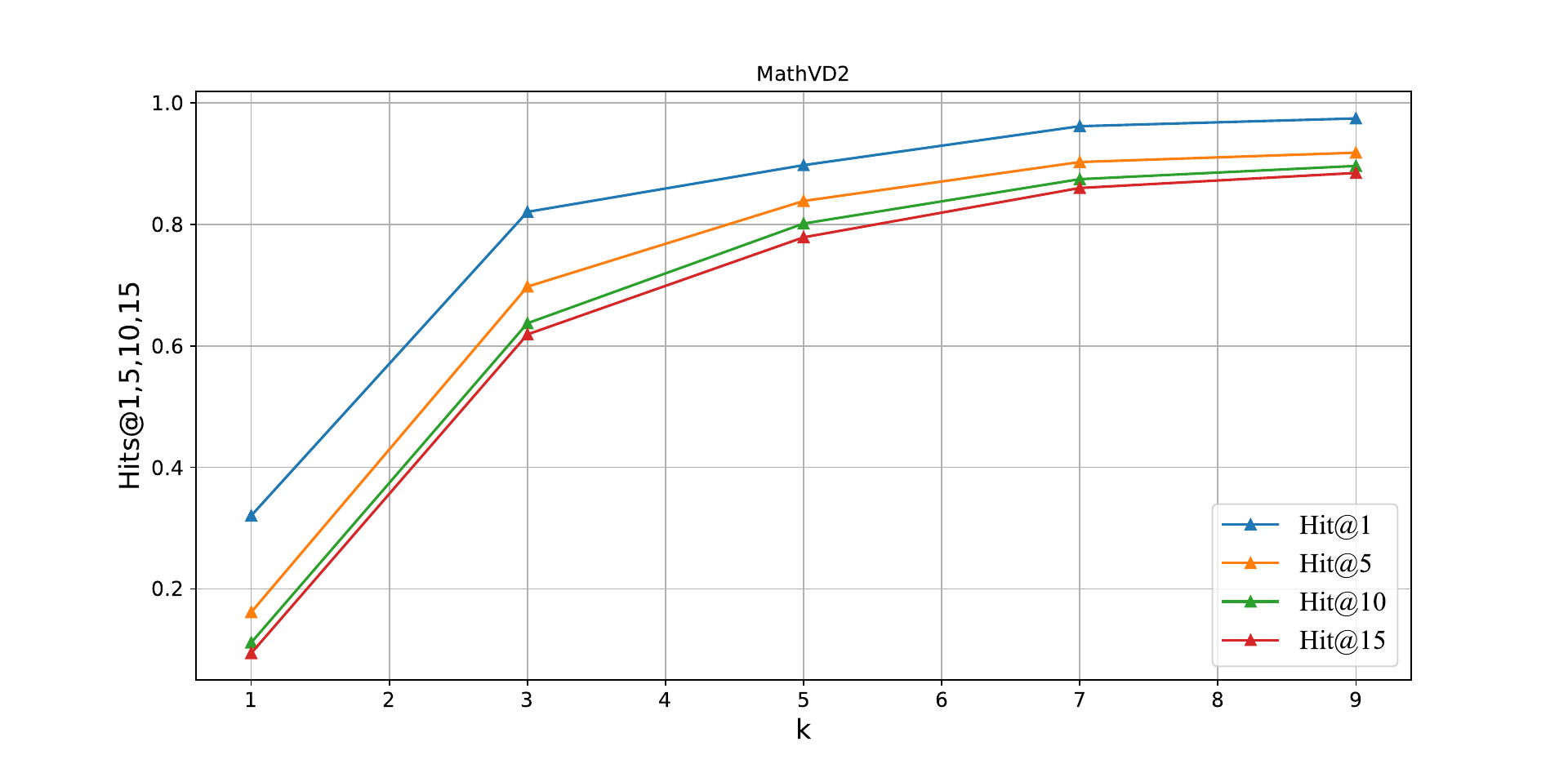}
\label{fig_MathVD2}
\end{subfigure}
}
\caption{The influence of hyperparameter $k$ on Hits@$q$ for MathVD1 and MathVD2.}    
\label{fig_MathVD}    
\end{figure*}
The results indicate that for query numbers $q$ = 5, 10, and 15, both MathVD models outperform all baseline models on Hits@$q$. This suggests that in our constructed VD, most retrieval queries based on cosine similarity are structurally related to the target entities in the KG, showing the effectiveness of our proposed embedding strategies for mathematical entities. Although our VDs are slightly surpassed by the classic TransE model when $q = 1$, their performance is notably superior at higher $q$ values. This is primarily because TransE is trained by predicting tail entities, which specifically optimizes for 1-hop reachability queries. Consequently, for a single query, TransE achieves a higher hit rate through supervised learning, but exhibits weaker generalization as the query number increases. In contrast, our model generalizes well across varying query numbers. Additionally, comparing our two VDs, it is found that MathVD2 performs slightly better for smaller $q$, while MathVD1 shows a little higher Hits@$q$ for larger $q$. 

There is one hyperparameter, namely the hop of reachability query denoted as $k$. $k$-hop reachability queries with different values of $k$ reflect varying degrees of association between entities. For $k = 1$, the 1-hop reachability query is equivalent to the common link prediction task. For $k > 1$, $k$-hop reachability queries could provide additional information about the relationship between two entities, such as multi-hop inference. Generally, entities that are reachable within smaller $k$ hops tend to exhibit stronger associations. Figure \ref{fig_MathVD} illustrates the influence of hyperparameter $k$ on Hits@$q$ for MathVD1 and MathVD2, with $q$ = 1, 5, 10 and 15. Initially, Hits@$q$ shows a sharp increase as $k$ increases from very small values; however, after $k$ increases to a certain threshold, the improvement becomes less significant. This indicates that most retrieved $k$-hop reachable entities can actually be accessed via smaller hops, resulting in increasing $k$ beyond this threshold having little effect on the hit rate. This not only highlights the insensitivity of the hit rate to larger values of the hyperparameter $k$, but also underscores the effectiveness of using cosine similarity to retrieve structurally related entities from our VDs.

\subsubsection{Math LLM reasoning capability}\label{secA4.3}

To assess the reasoning behavior of our designed Math LLM, we developed a framework for evaluating the mathematical capabilities of LLMs, based on an extensive literature review. Specifically, we rated the responses generated by the LLM on a scale from 1 to 5, with 5 representing the highest rating. The scoring criteria are as follows: a rating of 5 is assigned when the answer is nearly perfect; 4 is assigned when the model provides a mostly correct answer but with some issues, such as the use of unnecessarily complex methods; 3 is assigned when the model demonstrates correct reasoning but makes significant errors in execution; 2 is assigned when only a small part of the reasoning process is correct; and 1 is assigned when the output is largely irrelevant to the correct answer.

\begin{figure}[h]
\centering
\includegraphics[width=0.7\textwidth]{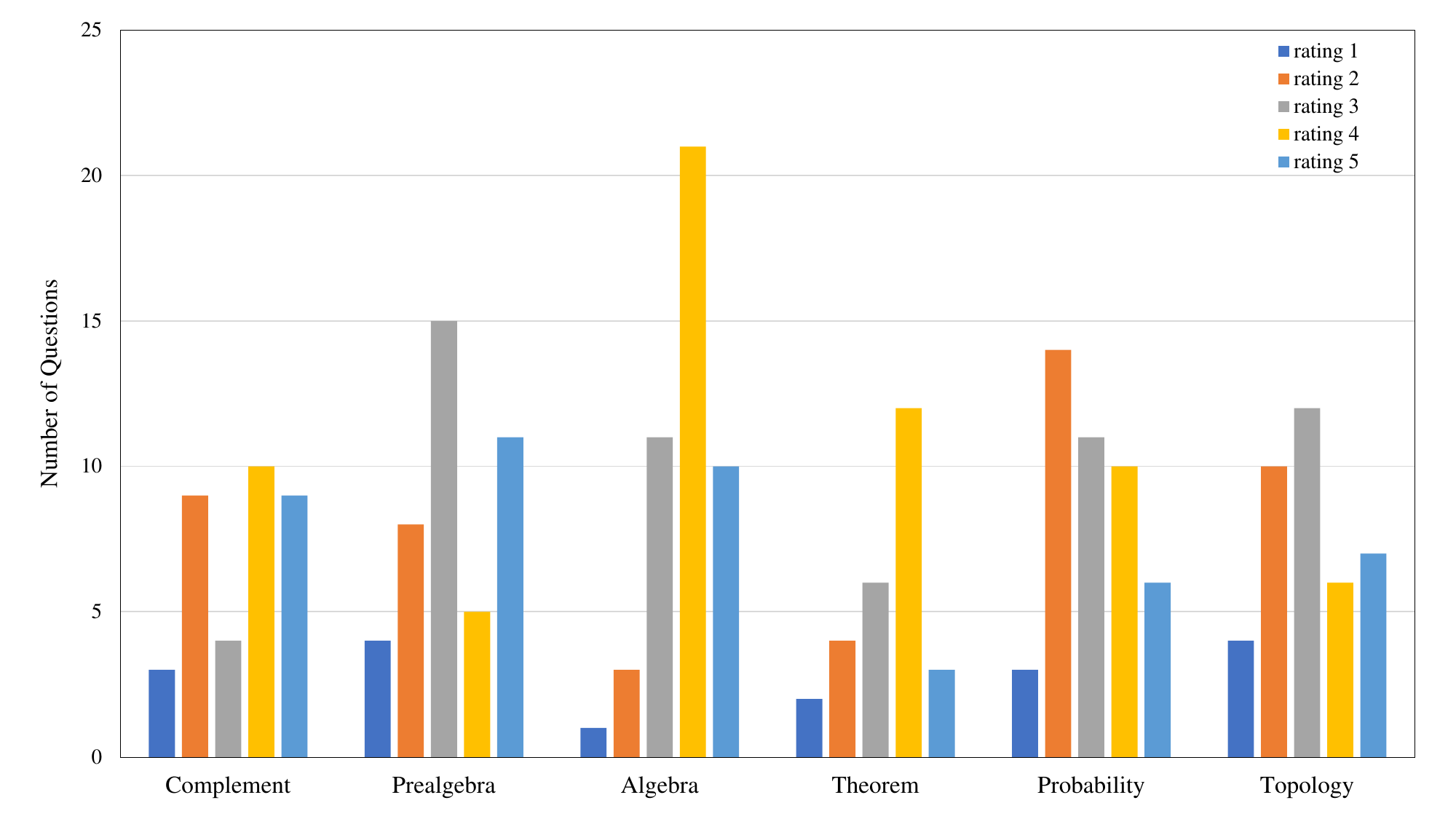}
\caption{Rating distribution in test questions.}\label{fig_LLMtest}
\end{figure}

In the experiments, 234 test questions were selected and refined from GHOSTS \citep{frieder2024mathematical}, categorized into six areas: Complement, Prealgebra, Algebra, Theorem, Probability, and Topology. Notably, the Complement category involves questions where parts of the proofs are intentionally omitted, requiring the LLM to fill in the gaps, which effectively aligns with the task of automatic knowledge completion by Math LLM. The evaluation was conducted using an online question-and-answer format, where the dataset question served as the prompt and the feedback was recorded in the output. Subsequently, we manually rated the answers generated by Math LLM according to the established framework. The sample size and average rating for each category of test questions are shown in Table \ref{tAA6}, and the rating distribution is illustrated in Fig. \ref{fig_LLMtest}. 

\begin{table*}[h]
\caption{The average evaluation rating of Math LLM.
}\label{tAA6}
\resizebox{\textwidth}{!}{
\begin{tabular}{lcccccc}
\hline
\textbf{Question category}  &\textbf{Complement} &\textbf{Prealgebra} & \textbf{Algebra} &\textbf{Theorem} &\textbf{Probability}  &\textbf{Topology}  \\
\hline
\textbf{Sample size} &35 &43 &46 &27 &44 &39\\
\hline
\textbf{Average rating} &3.3714                & 3.2558            & 3.7826                & 3.3704               & 3.0455             & 3.0513 \\
\hline
\end{tabular}
}
\end{table*}

The evaluation results indicate that Math LLM is capable of demonstrating correct reasoning and providing mostly correct answers in most cases, particularly for challenging problems involving theorem proving. This performance is commendable for a model with 7 billion parameters. Leveraging the various modules we have designed, Math LLM exhibits strong understanding, proof, and computational capabilities, thereby demonstrating its reliability for automatic knowledge completion.

\subsubsection{Internal retrieval relevance}
To assess the internal quality of our VD, we conducted human evaluations to quantify the precision of relevance in retrieving similar entities across both VDs. 
For any entity vector $e$, the top $t$ entities ranked by cosine similarity were retrieved as predictions of $t$ similar entities. If the retrieved entity is related to the target entity in terms of content, domain, or application, it is considered a correct similar prediction. Let $s$ denote the number of correct predictions for entity vector $e$. The precision $p$ of similar entity retrieval for entity vector $e$ is defined as follows:
\begin{align}
     p = \frac{s}{t}.
\end{align}

In the experiments, 50 random test samples were drawn from each VD based on the category proportions, including 25 Definition entities, 15 Theorem entities, and 10 Problem entities. These samples are denoted as $\{e_1^{(1)},\cdots,e_{50}^{(1)}\}$ for MathVD1, and $\{e_1^{(2)},\cdots,e_{50}^{(2)}\}$ for MathVD2 respectively. Similar entity retrieval was performed for each sample, with $t=10$. After manually evaluating the prediction results, the precision of similar entity retrieval for each sample was calculated, resulting in two series of precision data,

\begin{align}
P^{(1)}=\{p_1^{(1)},\cdots,p_{50}^{(1)}\}, P^{(2)}=\{p_1^{(2)},\cdots,p_{50}^{(2)}\}. 
\end{align}

\begin{figure}[h]
\centering
\includegraphics[width=0.7\textwidth]{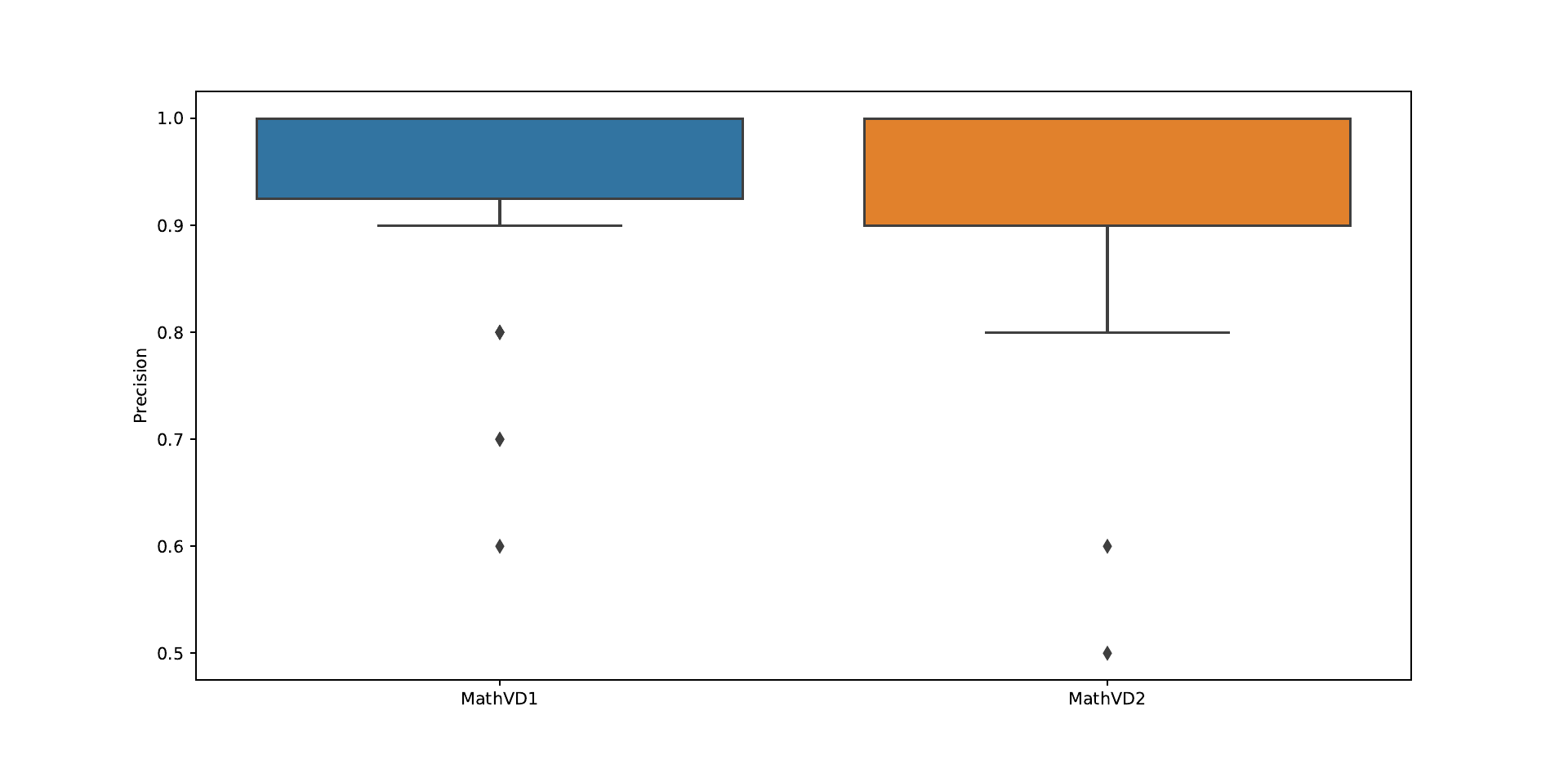}
\caption{The box plots of MathVD1 and MathVD2.}\label{fig8}
\end{figure}

Figure \ref{fig8} presents box plots of $P^{(1)}$ and $P^{(2)}$, visually illustrating the detailed distribution of two sets of precision data, both of which are concentrated around 1. A Kolmogorov-Smirnov (K-S) test was conducted on $P^{(1)}$ and $P^{(2)}$, indicating no significant difference in the distribution between the two sets of data. Additionally, Table \ref{t6} compares the average precision with the standard deviation (STDEV) of similar entity retrieval within MathVD1 and MathVD2. The precision of similar retrieval for all three types of entities exceeds 90\%, with MathVD1 exhibiting better performance in the retrieval of Definition and Problem entities, while performing equally well as MathVD2 in Theorem entity retrieval. The average precision of MathVD1 and MathVD2 are both around 95\%, with relatively small fluctuations. Therefore, MathVD1 and MathVD2 are both effective in capturing the relationships among internal entities, enabling the retrieval of relevant entities.

\begin{table}[h]
\caption{The comparison results of average precision for similar entity retrieval in MathVD1 and MathVD2.}\label{t6}
\begin{tabular*}{\textwidth}{@{\extracolsep\fill}ccccc@{\extracolsep\fill}}
\hline
\textbf{Sample type} &\textbf{Def}       & \textbf{Thm}   & \textbf{Prob}   & \textbf{All (STDEV)} \\
\hline
\textbf{MathVD1}        & 96.0\%    & 94.0\%  &97.0\%   & \textbf{95.2\%} (0.0953) \\
\textbf{MathVD2}         & 95.3\%              & 94.0\%    & 96.0\% &\textbf{94.8\%} (0.1035)      \\
\hline
\end{tabular*}
\end{table}

\subsubsection{External retrieval relevance}
To assess the external applicability of our VD, we conducted experiments on similar entity retrieval for external queries to evaluate their relevance.
First, the queries are embedded into vectors using SBERT, after which similarity retrieval in MathVD is performed, similar to the internal entity retrieval process. Tables \ref{t7} and \ref{t8} present the retrieved entity contents for the same external query using MathVD1 and MathVD2, respectively. For the query of \textit{``Expectation in Probability and Statistics"}, both MathVD1 and MathVD2 retrieve definitions and theorems related to \textit{``Expectation"}, while MathVD1 also retrieves relevant problems, assisting the query requester in understanding this mathematical concept from different perspectives. Therefore, for queries from external sources, both of our VDs are capable of retrieving various types of highly relevant entities, indicating their effectiveness in handling external queries.

\begin{table}[h]
\caption{The results of external query example in MathVD1.}\label{t7}
\begin{tabular}{ll}
\hline
\multicolumn{2}{l}{\textbf{External query: Expectation in Probability and Statistics}} \\
\hline
\multicolumn{2}{l}{\textbf{Retrieval results in MathVD1}}                                   \\
\hline
\multicolumn{1}{c}{\textbf{Rank}}    & \multicolumn{1}{c}{\textbf{Entity}}   \\
\hline
1                                    & \makecell[l]{Theorem: expectation of geometric distribution\\
Let $X$ be a discrete random variable with the geometric distribution with parameter $p$.\\ Then the expectation of $X$ is given by:$ E(X)=\frac{1-p}{p}$.}
                                    \\
\hline                                    
2                                    & \makecell[l]{Problem: expectation of waiting time\\
Let $X_0,X_1,X_2,\cdots$ be drawn $i.i.d.$ from $p(x)$, and $x\in\{1,2,3,\cdots,100\}$. \\Let N be the waiting time to the next occurrence of $X_0$. Compute $E(N)$.}
     \\
\hline
3                                    & \makecell[l]{Theorem: expectation of discrete uniform distribution\\ 
Let $X$ be a discrete random variable with the discrete uniform distribution with \\parameter $n$. Then the expectation of $X$ is given by: $ E(X)=\frac{n+1}{2}$.
}
                                    \\
\hline 
4                                    & \makecell[l]{Definition: expectation of random vector\\
Let $X_1,X_2,\cdots,X_n$  be random variables on a probability space $(\Omega ,\Sigma ,Pr)$. \\Let $X=(X_1,X_2,\cdots,X_n)$ be a random vector. Then the expected value of $X$, $E(X)$, \\is defined by: $E(X)=(E(X_1),E(X_2),\cdots,E(X_n))$.}
                                    \\
\hline 
5                                    & \makecell[l]{Theorem: sum of expectations of independent trials:\\
Let$ E_1,E_2,\cdots,E_n$ be a sequence of experiments whose outcomes are independent of \\each other. Let $X_1,X_2,\cdots,X_n$ be discrete random variables on $E_1,E_2,\cdots,E_n$ \\respectively. Let $E(X_j)$ denote the expectation of $X_j$ for $j\in\{1,2,\cdots,n\}$. Then \\we have, whenever both sides are defined: $ E({\textstyle \sum_{j=1}^{n}X_j})={\textstyle \sum_{j=1}^{n} E(X_j)}$. \\That is, the sum of the expectations equals the expectation of the sum.
}\\
\hline 
\end{tabular}
\end{table}

\begin{table}[h!]
\caption{The results of external query example in MathVD2.}\label{t8}
\begin{tabular}{ll}
\hline
\multicolumn{2}{l}{\textbf{External query: Expectation in Probability and Statistics}} \\
\hline
\multicolumn{2}{l}{\textbf{Retrieval results in MathVD2}}                                   \\
\hline
\multicolumn{1}{c}{\textbf{Rank}}    & \multicolumn{1}{c}{\textbf{Entity}}   \\
\hline
1                                    & \makecell[l]{Definition: expectation \\
The expectation of a random variable is the arithmetic mean of its value.
}
                                    \\
\hline                                    
2                                    & \makecell[l]{Theorem: expectation of binomial distribution\\
Let $X$ be a discrete random variable with the binomial distribution with parameters \\$n$ and $p$ for some $n\in \mathbb{N}$ and $0\le p\le 1$. Then the expectation of $X$ is given by: \\$E(X)=np$.}
     \\
\hline
3                                    & \makecell[l]{Theorem: expectation of geometric distribution\\
Let $X$ be a discrete random variable with the geometric distribution with parameter $p$.\\ Then the expectation of $X$ is given by:$ E(X)=\frac{1-p}{p}$.}
                                    \\
\hline 
4                                    & \makecell[l]{Theorem: sum of expectations of independent trials:\\
Let$ E_1,E_2,\cdots,E_n$ be a sequence of experiments whose outcomes are independent of \\each other. Let $X_1,X_2,\cdots,X_n$ be discrete random variables on $E_1,E_2,\cdots,E_n$ \\respectively. Let $E(X_j)$ denote the expectation of $X_j$ for $j\in\{1,2,\cdots,n\}$. Then \\we have, whenever both sides are defined: $ E({\textstyle \sum_{j=1}^{n}X_j})={\textstyle \sum_{j=1}^{n} E(X_j)}$. \\That is, the sum of the expectations equals the expectation of the sum.
}\\
\hline 
5                                    & \makecell[l]{Theorem: expectation of bernoulli distribution\\
Let $X$ be a discrete random variable with a Bernoulli distribution with parameter $p$. \\Then the expectation of $X$ is given by: $E(X)=p$.
}
                                    \\
\hline 
\end{tabular}
\end{table}

\section{Discussion}\label{sec8}
\subsection{Comparison with existing mathematical KGs}

There are significant advantages in AutoMathKG in terms of corpus sources, entity coverage and update methods, compared to existing mathematical KGs written in natural language, including MathGraph \citep{zhao2019mathgraph}, MathGloss \citep{horowitz2023mathgloss}, Math-KG \citep{wang2022math}, and NaturalProofs \citep{welleck2021naturalproofs}, as shown in Table \ref{t9}.

For corpus sources, AutoMathKG not only includes commonly covered web data and multiple high-quality textbooks, but also incorporates synthetic data and mathematical paper data that other mathematical graphs lack. The synthetic data augmented by LLM provides more detailed mathematical knowledge, while mathematical papers offer deeper domain-specific knowledge, giving AutoMathKG an advantage in richness and depth of math knowledge.
For entity coverage, AutoMathKG breaks the limitation of existing mathematical graphs focusing on only one aspect of mathematical concepts or problems. It views mathematics as a vast multidimensional network composed of definitions, theorems, and problems, fully exploring the relationships between them, and generating a directed graph that captures the interdependencies of mathematical knowledge.
For update methods, similar to existing mathematical graphs, AutoMathKG can integrate knowledge from different sources. The difference lies in the ability of AutoMathKG to automatically supplement incomplete knowledge and add it to the existing graph. Considering the abundance of incomplete knowledge in existing mathematical text corpora, AutoMathKG effectively enhances the utilization of existing corpus data and reduces human resources through LLM automatic updates.
\begin{table*}[h]
\caption{The comparison results between AutoMathKG and existing math KGs.}\label{t9}
\resizebox{\textwidth}{!}{
\begin{threeparttable} 
\begin{tabular}{lccccccccc}
\hline
\multirow{2}{*}{\textbf{KG}} & \multicolumn{5}{c}{\textbf{Corpus source}} & \multicolumn{3}{c}{\textbf{Entity coverage}} & \multicolumn{1}{c}{\textbf{Update method}} \\
& Online & Synthetic &Exercise &  \#Textbook &  \#Paper &Thm &Def &Prob &Automated \\
\hline                            
MathGraph         & -      & -     &\checkmark & -     & -    & \checkmark     & \checkmark      & \checkmark    & -     \\
MathGloss         & \checkmark      & -     & - & -     & -    & -     & \checkmark      & -    & -      \\
Math-KG           & \checkmark      & -     & - & -     & -    & \checkmark     & \checkmark      & -    & -      \\
NaturalProofs     & \checkmark      & -     & - & 2     & -    & \checkmark     & \checkmark      & -    & -       \\
AutoMathKG        & \checkmark      & \checkmark  & \checkmark   & 8     & 20   & \checkmark     & \checkmark      & \checkmark    & \checkmark    \\
\hline
\end{tabular}
\end{threeparttable}
}
\begin{tablenotes}
\item \footnotesize {\it Note}: \Checkmark and - denote the presence or absence of the item, respectively.
\end{tablenotes}
\end{table*}

\subsection{Ablation study for knowledge representation}
This paper proposes an innovative approach to knowledge representation for MathVD by leveraging the dependency relationships between entities within the AutoMathKG graph structure. To evaluate the contribution of reference information from incoming and outgoing entities in entity embeddings, we conducted an ablation study by removing this reference information to observe the resulting changes in performance for similar vector retrieval. Specifically, for each entity node, the descriptive sentences corresponding to the ``in\_refs" and ``out\_refs" attributes in Section 5 were removed, and only those corresponding to the ``title", ``field" and ``contents" attributes were used for vector representation, resulting in a VD variant, denoted as VD-no refs.

Table \ref{t10} shows a comparison of the performance in similar entity retrieval between the variant and the original MathVD using the first vector representation method proposed in Section \ref{sec5.1}, denoted as VD1-no refs and VD1-all, respectively. For the target Theorem entity \textit{``probability measure is subadditive"}, the top 5 similar entities retrieved in VD1-no refs are all intuitively related to the target entity in the field, with most of their entity titles containing \textit{``probability measure"} or \textit{``probability"}. In contrast, the first entity retrieved in VD1-all seems not similar to the target entity in terms of title, making it difficult for people from non-mathematical fields to identify the connection between them based on the title. However, upon examining the content of this entity, it is found to be a direct consequence of the target entity. This indicates that VD1-all can retrieve entities that are similar in essence to the target entity, not solely relying on linguistic similarity. Furthermore, another example in Table \ref{tA6} of Appendix \ref{secA5} shows that the cosine similarity between similar entities in VD1-all is higher overall, indicating denser similar entities in the vector space of VD1-all. The reason for this advantage is that VD1-all contains the interdependence between entities and stores it in the vector representation. These results demonstrate that the reference information provided by incoming and outgoing entities plays a crucial role in the vector representation of mathematical knowledge, enabling to retrieve essentially relevant entities. 

\begin{table}[h]
\caption{The results of ablation study.}\label{t10}
\begin{tabular*}{\textwidth}{@{\extracolsep\fill}cll@{\extracolsep\fill}}
\hline
\multicolumn{3}{l}{\textbf{\makecell[l]{Theorem: probability measure is subadditive\\Let $(\Omega ,\Sigma ,Pr)$ be a probability space. Then $Pr$ is a subadditive function.}}}              \\
\hline\hline
\multicolumn{3}{l}{\textbf{Retrieval results in VD1-no refs}}                                              \\
\hline
\textbf{Rank} & \textbf{Score} & \textbf{Entity}                                      \\
1             & 0.7632         & Theorem:elementary properties of probability measure \\
2             & 0.7362         & Definition:probability measure/definition 1          \\
3             & 0.7324         & Definition:probability function                      \\
4             & 0.6869         & Definition:probability measure/definition 2          \\
5             & 0.6343         & Probability measure is monotone                      \\
\hline\hline
\multicolumn{3}{l}{\textbf{Retrieval results in VD1-all}}                                                  \\
\hline
\textbf{Rank} & \textbf{Score} & \textbf{Entity}                                      \\
1             & 0.7632         & \textbf{Theorem: boole's inequality}   \\
2             & 0.7237         & Measure is subadditive                               \\
3             & 0.6980          & Definition:probability function                      \\
4             & 0.6960         & Definition:probability measure/definition 1          \\
5             & 0.6827         & Theorem:elementary properties of probability measure \\
\hline
\end{tabular*}
\end{table}

\subsection{Case study for automatic updates}
\subsubsection{Knowledge completion}
In order to achieve automatic knowledge completion, this paper utilizes the designed Math LLM to supplement incomplete proofs and solutions for new theorems and problems. Table \ref{t11} shows the supplementation effects of theorem proofs and problem solutions by Math LLM, where the incomplete entities are packaged as \textit{Question} sent to Math LLM, and the supplementation knowledge is output as \textit{Answer}.

\begin{table}[h]
\caption{The results of the case study on automatic knowledge completion.}\label{t11}
\begin{tabular}{ll}
\hline
\multicolumn{2}{l}{\textbf{Theorem proof}} \\
\hline
Question                & \makecell[l]{Let $f$ have a continuous derivative on $[a,b]$. We want to show that ${f}'$ satisfies a \\Lipschitz condition on $[a,b]$. By [MISSING], $f(x) - f(y) = {f}'(x_i)(x-y)$  for some \\$x_i\in (y,x)$. But since ${f}'$ is continuous and $[a,b]$ is compact, then ${f}'$ is bounded in \\that interval, say by $C$. Thus taking absolute values yields $|f(x)- f(y)|\le C|x-y|$.\\
Can you name the result used in [MISSING]?
}\\
\hline
Answer                & \makecell[l]{The answer is the Mean Value Theorem. The missing part in the text is the Mean \\Value Theorem. The Mean Value Theorem states that for a continuous function $f$ \\on the interval $[a,b]$, there exists a point $c$ in $(a,b)$ such that $f(b)-f(a) ={f}'(c)$\\$(b-a)$. This theorem is used to show that f satisfies a Lipschitz condition on $[a,b]$.
}\\
\hline\hline
\multicolumn{2}{l}{\textbf{Problem solution}} \\
\hline
Question                & \makecell[l]{What is the smallest positive multiple of 6 that is also a multiple of 8?
Solution: \\To find a positive number that is both a multiple of 6 and a multiple of 8, we'll \\list out the multiples of 6, starting with 6, and check to see if any of them are also \\multiples of 8. 6 does not divide 8, so it isn't a multiple of 8. 12 and 18 are also not\\ evenly divisible by 8. [MISSING]. Fill in the missing steps.
}\\
\hline
Answer                & \makecell[l]{Sure, here is the answer to the question:
To find the smallest positive multiple of \\6 that is also a multiple of 8, we'll list out the multiples of 6, starting with 6, and \\check to see if any of them are also multiples of 8. 6 does not divide 8, so it isn't a \\multiple of 8. 12 and 18 are also not evenly divisible by 8. The next multiple of \\6 is 24, and it is also a multiple of 8. Therefore, the smallest positive multiple of 6 \\that is also a multiple of 8 is 24.
}\\
\hline
\end{tabular}
\end{table}

\subsubsection{Knowledge fusion}
In order to achieve automatic knowledge fusion, this paper utilizes VDs and LLM to update entities, ensuring that entities and relationships from different sources are added to the Existing KG without redundancy or omission. Table \ref{t12} presents an example of Theorem entity fusion from different sources. The Theorem entity \textit{``Mean Value Theorem"} from a textbook is a necessary condition for the Theorem entity \textit{``Carathéodory's Theorem (Analysis)"} from ProofWiki. During the update of AutoMathKG, the two entities were successfully merged, and the reference relationships associated with the \textit{``Mean Value Theorem"} entity were added to the \textit{``Carathéodory's Theorem (Analysis)"} entity.

\begin{table}[ht]
\caption{The results of the case study on automatic knowledge fusion.}\label{t12}
\begin{tabular}{ll}
\hline
\textbf{Source} & \textbf{ProofWiki}                \\
\hline
Theorem         & Carathéodory's Theorem (Analysis)\\
\hline
Contents         & \makecell[l]{Let $I\subseteq \mathbb{R}$. Let $c\in I$ be an interior point of $I$. Let $f:I \to \mathbb{R} $ be a real function. \\Then $f$ is differentiable at $c$ if and only if: 
There exists a real function $\varphi: I \to \mathbb{R} $ \\that is continuous at $c$ and satisfies: \\(1): $\forall x\in I,f(x)-f(c)=\varphi(x)(x-c)$ (2): $\varphi(c)={f}'(c)$.
}\\
\hline\hline
\textbf{Source} & \textbf{Textbook}                \\
\hline
Theorem         & Mean Value Theorem\\
\hline
Contents         & \makecell[l]{If $f$ is continuous on the closed interval $[a,b]$ and differentiable on the open interval\\ $(a,b)$, then, ${f}'(c)=\frac{f(b)-f(a)}{b-a}$, for some $c$ in $(a,b)$.}\\
\hline

\end{tabular}
\end{table}

\subsection{Limitation}
One limitation of our work is the choice of LLM during ICL. To save computational resources, we only used a small-scale model, Llama-2-7b, while larger language models may perform better. Another limitation is the trade-off between corpus richness and mathematical knowledge specialization. To obtain accurate mathematical knowledge, the corpus collected in this paper focuses on formal texts and does not include informal corpora, such as the discussion dialogues in mathematical forums, which may provide different types of mathematical knowledge.

\section{Conclusion}\label{sec9}
This paper proposes AutoMathKG, a novel high-quality, wide-coverage, and multi-dimensional knowledge graph in the field of mathematics that can be automatically updated. AutoMathKG integrates math knowledge from various domains and levels in natural language, establishing connections among the three major dimensions of mathematics: definitions, theorems, and problems, all augmented by LLMs. In order to retrieve similar entities, MathVD is created as a VD of AutoMathKG, with two strategies for entity embeddings using SBERT. In order to automatically update, two mechanisms are proposed. One is for knowledge completion through interacting with Math LLM, a specialized LLM designed to supplement missing proofs and solutions. Another is for knowledge fusion across different sources, through fuzzy search in MathVD and the determination of update approaches by LLM.
Comprehensive experimental results demonstrate the advanced performance of AutoMathKG, MathVD, and Math LLM, as well as the effectiveness and applicability of the automatic update method. AutoMathKG system opens up new possibilities for applying AI techniques in the acquisition of mathematical knowledge and the enhancement of mathematical reasoning, making a significant contribution to the field of AI for mathematics.

In future work, we plan to use more advanced models beyond Llama-2-7b for data augmentation. Moreover, we intend to explore a broader range of mathematical corpora, such as math forums, to increase the diversity of mathematical knowledge.

\backmatter

\bmhead{Acknowledgements}
This research was supported by Key Lab of Random Complex Structures and Data Science, Chinese Academy of Sciences (2008DP173182); National Key R\&D Program of China (2021ZD0111204).

\section*{Declarations}

\begin{itemize}
\item \textbf{Author contribution} All authors contributed to this research. Data collection, model implementation, and experimental evaluation were conducted by R.B., Y.G., and Z.Y. B.C. formulated the research idea, coordinated the efforts, and supervised the overall process. R.B. drafted the manuscript, and all authors reviewed and commented on its previous versions.
\item \textbf{Funding} This research was supported by Key Lab of Random Complex Structures and Data Science, Chinese Academy of Sciences (2008DP173182); National Key R\&D Program of China (2021ZD0111204).
\item \textbf{Conflict of interest} The authors declare that they have no conflict of interest.
\item \textbf{Ethics approval and consent to participate} Not applicable.
\item \textbf{Consent for publication} Not applicable.
\item \textbf{Data availability} The data collected to construct the knowledge graph can be accessed through the provided links or citations in this article. The constructed knowledge graph will be made available on request.

\end{itemize}

\noindent


\begin{appendices}

\section{More information for corpus}\label{secA1}
\vspace*{12pt}
Table \ref{tA1} presents the title information of the eight textbooks analyzed in this study. The first six textbooks were found through an online search for open-source textbooks, while the last two were provided by the NaturalProofs dataset. Table \ref{tA2} shows the subject information of the arXiv papers collected in this study.
\begin{table}[h]
\caption{The title information of selected textbooks.}\label{tA1}
\begin{tabular*}{\textwidth}{@{\extracolsep\fill}l@{\extracolsep\fill}}
\hline
\textbf{Textbook titles}\\  
\hline
Linear Transformations on Vector Spaces \\
Elementary Differential\\
Basic Analysis: Introduction to Real Analysis\\
Ossifrage and Algebra\\
Multivariable and Vector Calculus\\
An Introduction to Group Theory\\
Introduction to Real Analysis\\
Elementary Number Theory: Primes, Congruences, and Secrets  \\
\hline
\end{tabular*}
\end{table}

\begin{table}[h]
\caption{The subject information of selected papers from arXiv.}\label{tA2}
\begin{tabular*}{\textwidth}{@{\extracolsep\fill}l@{\extracolsep\fill}}
\hline
\textbf{Paper subjects}\\
\hline
\makecell[l]{math.AG, math.AT, math.AP, math.CT, math.CA, math.CO, math.AC,   math.CV, math.DG, \\math.PR, math.ST}\\
\hline
\end{tabular*}
\end{table}

\section{Instructions of entity storage format}\label{secA2}
\vspace*{12pt}
Table \ref{tA3} details the entity storage format, including the storage type and structure of all attributes as well as the applicable entity types.

\begin{table}[h]
\caption{The attributes stored in entities of AutoMathKG.}\label{tA3}
\begin{tabular*}{\textwidth}{@{\extracolsep\fill}llll@{\extracolsep\fill}}
\hline
\textbf{Attribute} & \textbf{Storage type} & \textbf{Storage structure} & \textbf{Entity} \\
\hline
Id                   & Int           & Number                     & Thm, Def, Prob  \\
\hline
Type                 & String        & Theorem/definition/problem & Thm, Def, Prob  \\
\hline
Label                & String        & Entity label               & Thm, Def, Prob  \\
\hline
Title                & String        & Entity title               & Thm, Def, Prob  \\
\hline
Field                & String        & \makecell[l]{Algebra/geometry/analysis/\\probability and statistics/\\applied mathematics/$\dots$}               & Thm, Def, Prob  \\   
\hline
Contents             & List        & [content1, content2, $\cdots$]               & Thm, Def, Prob  \\
\hline
Bodylist             & \makecell[l]{List nested\\dictionary}        & \makecell[l]{[\{``description”: content1,\\
``action”: action1\},$\dots$]}               & Thm, Def  \\
\hline
Refs                 & List        & [reference1, reference2, $\dots$]               & Thm, Def, Prob  \\
\hline
References\_tactics   & Dictionary        & \makecell[l]{\{``reference1”: tactic1,\\
``reference2”: tactic2,$\dots$\}}               & Thm, Def, Prob  \\
\hline
Source               & String        & \makecell[l]{ProofWiki/textbook/arXiv/TheoremQA}               & Thm, Def, Prob  \\
\hline
Proofs               & \makecell[l]{List nested\\dictionary}        & \makecell[l]{[\{``contents”: [$\cdots$],\\``refs”: [$\cdots$],\\``bodylist”: [$\cdots$],\\``references\_tactics” : $\{\cdots\}$\},$\cdots$]}               & Thm  \\
\hline
Solutions            & \makecell[l]{List nested\\dictionary}        & \makecell[l]{[\{``contents”: [$\cdots$],\\``refs”: [$\cdots$],\\``bodylist”: [],\\``references\_tactics” : $\{\cdots\}$\},$\dots$]}               & Prob  \\
\hline
In\_refs              & Dictionary        & \makecell[l]{\{``in\_reference1”: tactic1,\\``in\_reference2”: tactic2,$\dots$\}}               & Thm, Def, Prob  \\
\hline
In\_ref\_ids           & List       & [in\_id1, in\_id2,$\dots$]               & Thm, Def, Prob  \\
\hline
Out\_refs             & Dictionary        & \makecell[l]{\{``out\_reference1”: tactic1,\\``out\_reference2”: tactic2,$\dots$\}}               & Thm, Def, Prob  \\
\hline
Out\_ref\_ids          & List        & [out\_id1, out\_id2,$\dots$]               & Thm, Def, Prob  \\
\hline
\end{tabular*}
\end{table}

\section{In-context learning templates}\label{secA3}
\vspace*{12pt}
Table \ref{tA4} presents the prompt templates used by LLM during ICL for enhancing entity attributes, such as ``title", ``field", and ``bodylist", for different types of entities. Table \ref{tA5} displays the prompt templates used in the automatic knowledge fusion process through ICL, including templates for Steps 4 and 5.

\begin{table}[h]
\caption{The templates of entity augmentation.}\label{tA4}
\begin{tabular}{lll}
\hline 
\textbf{Attribute}                 & \textbf{Entity}            & \textbf{Template}   \\
\hline 
Title              & \makecell[l]{Def\\ Thm \\Prob}                      & \makecell[l]{Your task is to generate a title that can\\ summarize the content of the given math \\definition/theorem/problem.}  \\
                                    
\hline 
Field              & \makecell[l]{Def\\ Thm \\Prob}                       & \makecell[l]{Your task is to identify the  most relevant \\mathematical field for the given math \\definition/theorem/problem from the \\following choices: ``algebra", ``geometry", \\``analysis", ``logic", ``probability and statistics", \\``applied mathematics", ``foundations of \\mathematics". Choose one from them.} \\                                    
\hline 
Bodylist           &  \makecell[l]{Def\\Thm\\(including proof)}                       & \makecell[l]{Your task is to label each   element in the \\given content list of a math definition/theorem/\\theorem proof, in order to determine the role of \\each element. The labels of mathematical roles \\are: ``premise", ``assumption", ``lemma", ``corollary", \\``definition", ``conclusion", ``deduction", ``calculation", \\``enumeration". Choose only the most relative \\one from them when you label.}     \\
\hline 
Refs                                 & \makecell[l]{Prob\\(including solution)} & \makecell[l]{Your task is to identify the mathematical \\definitions or theorems referenced in the given \\math problem/problem solution. When there \\is at least one reference, output each reference \\in the following format: ``definition:" or ``theorem:" \\along with the original reference. When there is \\no reference, output an empty list. }                                                                                                         \\
\hline 
Referenecs\_tactics         & \makecell[l]{Def\\ Thm\\(including proof)\\Prob\\(including solution)}                      & \makecell[l]{Your task is to label each reference shown \\in the given content list of a math definition/\\theorem/theorem proof/problem/problem \\solution, in order to determine the roles of each \\reference. The labels of mathematical roles are: \\``premise", ``assumption", ``lemma", ``corollary", \\``definition", ``conclusion", ``deduction", ``calculation", \\``enumeration". Choose only the most relative one \\from them when you label.} \\                             \hline                                      
\end{tabular}
\end{table}

\begin{table}[h]
\caption{The templates of automatic knowledge fusion in Steps 4 and 5.}\label{tA5}
\begin{tabular*}{\textwidth}{@{\extracolsep\fill}ll@{\extracolsep\fill}}
\hline
\textbf{Automatic knowledge fusion} & \textbf{Template} \\
\hline
Step 4     & \makecell[l]{Your task is to decide if the first math theorem mean the same \\thing as the second theorem. If they have the same meaning,\\ you should answer ``yes". Otherwise, you answer ``no".} \\
\hline
Step 5    & \makecell[l]{Your task is to decide which of the candidate theorems mean \\the same thing as the new theorem. You should choose one \\candidate and output its id number as your answer.}\\ \hline
\end{tabular*}
\end{table}

\section{Math LLM training details}\label{secA4}
\vspace*{12pt}
\subsection{Training datasets}\label{secA4.1}
The gemma-7b-it model was chosen as the base model for the construction of MathLLM. 
Firstly, the model was fine-tuned using LORA on the MathInstruct \citep{yue2023mammoth} dataset, which contains various types of mathematical problem sets, enabling the model to acquire foundational problem-solving capabilities. Furthermore, the application adapter was trained on the ``Microsoft/orca-math-word-problems-200k" \citep{mitra2024orca} dataset, which includes a diverse range of application problems. The sympy adapter was trained on the GSM8K-sympy-v2 \citep{cobbe2021training} dataset, enhancing its ability to solve problems using the SymPy package. The proof adapter was trained on the NaturalProofs \citep{welleck2021naturalproofs} dataset, which provides a comprehensive collection of definitions, theories, and their corresponding proofs.
 
\subsection{Training hyperparameters}\label{secA4.2}
For all training processes, including base model fine-tuning and task adapter training, we employed fully shared data parallelism (FSDP) with the configuration parameters: node=1 and proc\_per\_node=4 in the PyTorch framework. The optimizer used was paged\_adaw\_32bit. The gemma-7b-it model was loaded in torch.bf16 format. The maximum sequence length was set to 1024. The LoRA alpha was configured to 16, LoRA rank to 8, and LoRA dropout to 0.1. The learning rate was set to 2e-4, with a weight decay of 0.001. The training batch size was set to 4, with fine-tuning performed for 50 epochs and adapters trained for 25 epochs.

\section{More examples for study}\label{secA5}
\begin{figure}[h]
\centering
\includegraphics[width=0.7\textwidth]{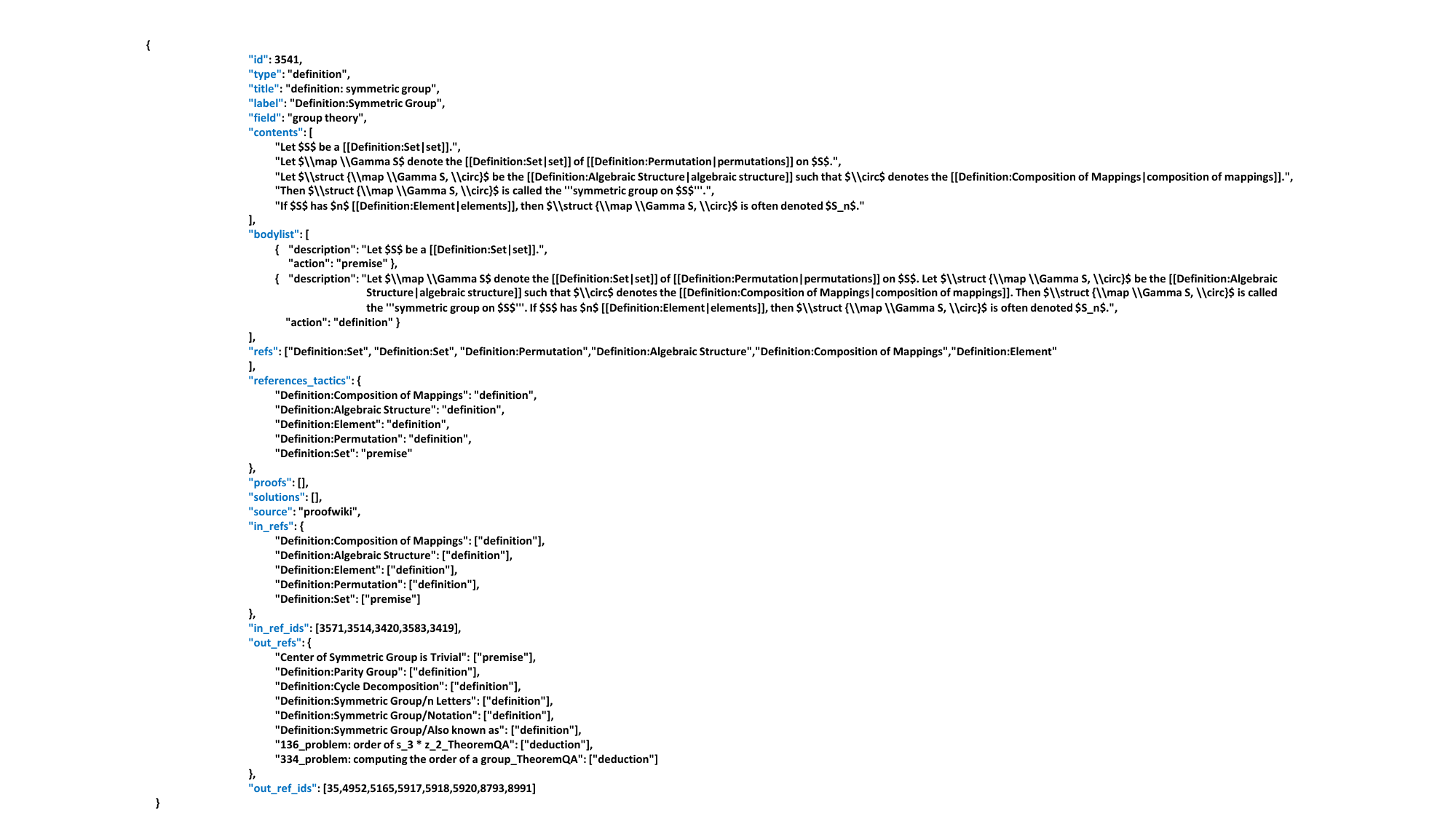}
\caption{An instance of a Definition entity in AutoMathKG.}\label{fig9}
\end{figure}

\vspace*{12pt}
Figure \ref{fig9} presents an instance of a Definition entity stored in JSON format. Table \ref{tA6} provides another example of similar entity retrieval for the ablation study.

\begin{table}[b]
\caption{The result of another example in ablation study.}\label{tA6}
\begin{tabular*}{\textwidth}{@{\extracolsep\fill}cll@{\extracolsep\fill}}
\hline
\multicolumn{3}{l}{\textbf{\makecell[l]{Theorem: union is associative\\$A\cup(B\cup C)=(A\cup B)\cup C$}}}              \\
\hline\hline
\multicolumn{3}{l}{\textbf{Retrieval results in VD1-no refs}}                                              \\
\hline
\textbf{Rank} & \textbf{Score} & \textbf{Entity}                                      \\
1             & 0.7341         & Theorem:intersection is associative \\
2             & 0.7317         & Theorem: union is commutative          \\
3             & 0.7298         & Theorem: union is idempotent                      \\
4             & 0.7090         & Theorem: union distributes over intersection          \\
5             & 0.7028         & Theorem: cartesian product distributes over union                      \\
\hline\hline
\multicolumn{3}{l}{\textbf{Retrieval results in VD1-all}}                                                  \\
\hline
\textbf{Rank} & \textbf{Score} & \textbf{Entity}                                      \\
1             & 0.8977         & Theorem: union is commutative   \\
2             & 0.8300         & Theorem: symmetric difference is associative                               \\
3             & 0.8217          & Theorem: set intersection is associative                      \\
4             & 0.7996          & Theorem: union as symmetric difference with intersection          \\
5             & 0.7808         & Theorem: union with complement \\
\hline
\end{tabular*}
\end{table}




\end{appendices}



\end{document}